\documentclass{article}

    \PassOptionsToPackage{numbers, compress}{natbib}
    \usepackage[final]{neurips_2025}

\usepackage[utf8]{inputenc} % allow utf-8 input
\usepackage[T1]{fontenc}    % use 8-bit T1 fonts
\usepackage{url}            % simple URL typesetting
\usepackage{booktabs}       % professional-quality tables
\usepackage{amsfonts}       % blackboard math symbols
\usepackage{nicefrac}       % compact symbols for 1/2, etc.
\usepackage{microtype}      % microtypography
\usepackage{xcolor}         % colors
\usepackage{amsmath}

\usepackage{amsthm}
\theoremstyle{plain}

\theoremstyle{definition}

\theoremstyle{remark}

\usepackage{multirow}
\usepackage{amsthm,amssymb,lipsum}
\usepackage{times}
\usepackage{mathrsfs}
\usepackage{enumerate}
\usepackage{colortbl}
\usepackage{bbding}
\usepackage{pifont}
\usepackage{amsmath}
\usepackage{wasysym}
\usepackage{textcomp}
\usepackage{microtype}      % microtypography
\usepackage[bottom]{footmisc}

\usepackage{color}
\usepackage{tocloft}
\usepackage{caption}
\usepackage{float}
\usepackage{afterpage}
\usepackage{xcolor}
\usepackage{graphicx}
\usepackage{booktabs}
\usepackage{multicol}
\usepackage{xspace}
\usepackage{times}
\usepackage{subfig}
\usepackage{wrapfig}
\usepackage{enumitem}

\def\eg{{\it{e.g.}}}

\def\ie{{\it{i.e.}}}

\def\sofar{{\scshape SoFar}\xspace}
\def\ours{{\scshape SoFar}\xspace}

\definecolor{drp-blue}{HTML}{1f77b4}
\definecolor{pretty-blue}{RGB}{0, 113, 188}
\definecolor{kaiming-green}{RGB}{57,181,74} % kaiming green
\definecolor{mypurple}{RGB}{55,0,168} % kaiming green
\definecolor{icmlblue}{rgb}{0,0.08,0.45} % ICML Blue
\definecolor{mygreen}{HTML}{4FC978}
\definecolor{linecolor1}{RGB}{246, 248, 239}
\definecolor{linecolor2}{RGB}{230, 234, 217}
\definecolor{linecolor3}{RGB}{211, 222, 190}
\definecolor{reconcolor}{HTML}{412F8A}
\definecolor{runpei-orange}{HTML}{F35F27}
\definecolor{runpei_blue}{HTML}{14294B}
\definecolor{datacolor}{HTML}{0009BF}
\definecolor{vitcolor}{HTML}{fc8e62}
\definecolor{cvprblue}{rgb}{0.21,0.49,0.74}
\definecolor{myblue}{rgb}{.39,.58,.93}

\newcommand{\worldwideweb}{\raisebox{-1.5pt}{\includegraphics[height=1.05em]{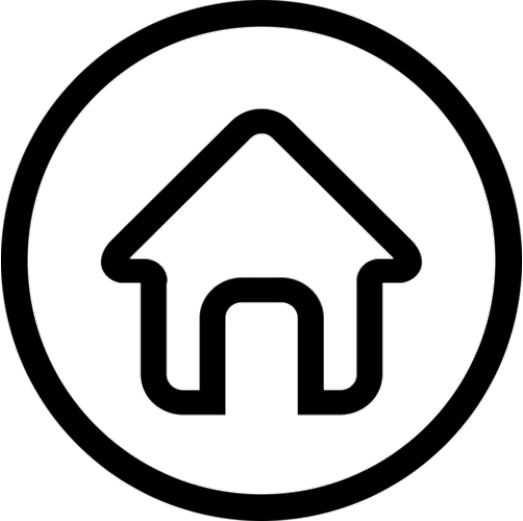}}\xspace}
\newcommand{\github}{\raisebox{-1.5pt}{\includegraphics[height=1.05em]{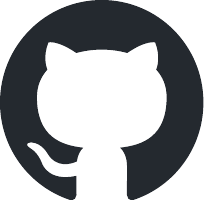}}\xspace}
\newcommand{\huggingface}{\raisebox{-1.5pt}{\includegraphics[height=1.05em]{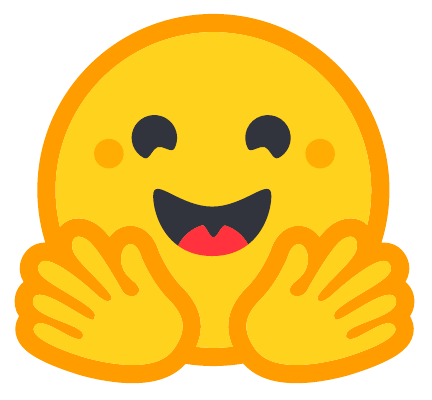}}\xspace}

\usepackage[
    bookmarks=true,
    pagebackref,
    breaklinks,
    colorlinks,
    linkcolor=blue, 
    urlcolor=blue,
    citecolor=runpei-orange,
]{hyperref}
\usepackage[capitalize]{cleveref}

\title{\sofar: Language-Grounded Orientation Bridges Spatial Reasoning and Object Manipulation}

\author{
  \makebox[0pt][c]{
  \textbf{
  Zekun Qi$^{13}$\thanks{Equal contribution. $^\ddagger$Corresponding author.} \quad 
  Wenyao Zhang$^{237*}$ \quad 
  Yufei Ding$^{34*}$ \quad 
  Runpei Dong$^{5}$ \quad 
  Xinqiang Yu$^{3}$
  }
  }
  \\
  \makebox[0pt][c]{
  \textbf{
  Jingwen Li$^{4}$\quad
  Lingyun Xu$^{4}$\quad
  Baoyu Li$^{5}$\quad
  Xialin He$^{5}$\quad
  Guofan Fan$^{1}$\quad
  Jiazhao Zhang$^{3}$\quad
  Jiawei He$^{3}$
  }
  }
  \\
  \makebox[0pt][c]{
  \textbf{
  Jiayuan Gu$^{6}$\quad
  Xin Jin$^{7}$\quad
  Kaisheng Ma$^{1}$\quad
  Zhizheng Zhang$^{3\ddagger}$\quad
  He Wang$^{34\ddagger}$\quad
  Li Yi$^{18\ddagger}$
  }
  }
  \vspace{4pt}
  \\
  $^{1}$Tsinghua University\quad
  $^{2}$Shanghai Jiao Tong University\quad
  $^{3}$Galbot\quad
  $^{4}$Peking University\quad
  $^{5}$UIUC\\
  $^{6}$ShanghaiTech University\quad
  $^{7}$Eastern Institute of Technology\quad
  $^{8}$Shanghai Qi Zhi Institute
  \vspace{5pt}
  \\
  {\worldwideweb \href{https://qizekun.github.io/sofar/}{{\text{Project Page}}}} \quad \quad {\github \href{https://github.com/qizekun/SoFar}{{\text{GitHub Code}}}} \quad \quad {\huggingface \href{https://huggingface.co/collections/qizekun/sofar-67b511129d3146d28cea9920}{{\text{HuggingFace}}}}
  \\
}

\begin{document}

\maketitle

\begin{center}
    \centering
    \vspace{-2.0em}
    \captionsetup{type=figure}
    \includegraphics[width=1\linewidth]{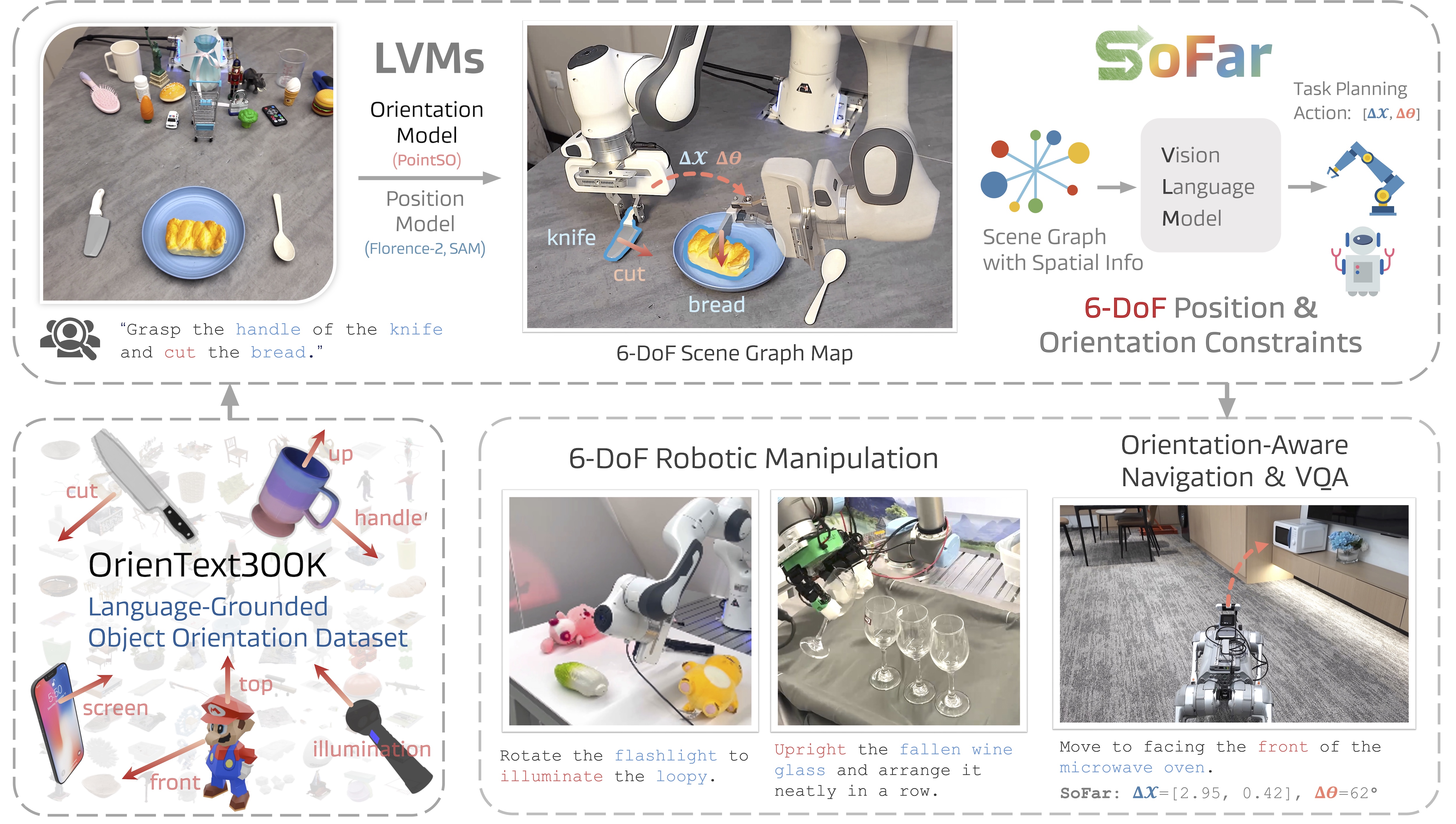}
    \vspace{-1.5em}
    \caption{
    We introduce the concept of \textit{Semantic Orientation}, which refers to natural language-grounded object orientations, such as the ``cutting'' direction of a knife or the ``handle'' direction of a cup. To support this, we construct OrienText300K, a large-scale object-text-orientation pairs dataset.
    }
    \label{fig:teaser}
\end{center}

\begin{abstract}
While spatial reasoning has made progress in object localization relationships, it often overlooks object orientation—a key factor in 6-DoF fine-grained manipulation. Traditional pose representations rely on pre-defined frames or templates, limiting generalization and semantic grounding. In this paper, we introduce the concept of semantic orientation, which defines object orientations using natural language in a reference-frame-free manner (\eg, the ``plug-in'' direction of a USB or the ``handle'' direction of a cup). To support this, we construct OrienText300K, a large-scale dataset of 3D objects annotated with semantic orientations, and develop PointSO, a general model for zero-shot semantic orientation prediction. By integrating semantic orientation into VLM agents, our \sofar~framework enables 6-DoF spatial reasoning and generates robotic actions. Extensive experiments demonstrated the effectiveness and generalization of our \sofar, \eg, zero-shot 48.7\% successful rate on Open6DOR and zero-shot 74.9\% successful rate on SIMPLER-Env.
\end{abstract}

% \keywords{Robotic Manipulation, Spatial Reasoning, Object Orientation} 
\section{Introduction}\label{sec:intro}

We observe that current VLMs struggle with understanding object \textbf{orientation}, making them insufficient for 6-DoF robot manipulation planning.
Consider some everyday scenarios: cutting bread in half with a knife, righting a tilted wine glass, or plugging a cord into a power strip. 
Previous approaches~\cite{SpatialVLM24,SpatialRGPT24,SpatialBot24} primarily focused on understanding ``\textit{where are the knife and wine glass}'' while ignoring their orientations—such as the ``blade direction'' of the knife and the ``up direction'' of the glass. This oversight makes it challenging to accomplish these 6-DoF manipulation tasks.

More importantly, different orientations of an object hold varying semantic significance. The capability of connecting specific orientations to their semantic meanings is essential for language-guided robot manipulations. For example, inserting a pen into a pen holder requires aligning the pen tip with the direction of the pen holder's opening; righting a wine glass necessitates aligning the glass's top with the z-axis in the world coordinate frame; and plugging into a power strip involves understanding the ``insertion'' direction, which is perpendicular to the power strip's surface.
However, translating a specific language description into a desired orientation is challenging for existing VLMs.

To move forward, we introduce \textit{language-grounded orientation that bridges spatial reasoning and object manipulation}, characterized by the following:
\begin{itemize}[leftmargin=1.5em]
\vspace{-3pt}
\item \textbf{From Position Awareness to Orientation Awareness.}
While prior works~\cite{SpatialVLM24,SpatialRGPT24,SpatialBot24} emphasize position relationship, orientation understanding is equally critical for defining the full 6-DoF of object pose or end-effector poses~\cite{OV6DOFPose24,FoundationPose24,PoseCNN18,ManipLLM24}. Orientation awareness involves understanding object orientations and their relationships in the open world, enabling robots to complete tasks requiring precise alignment and rearrangement.
% \vspace{-2pt}
\item \textbf{From Orientation to Semantic Orientation.}
Traditional orientation, defined relative to a base frame or template model~\cite{OnePose22,MegaPose22,FoundationPose24,OV6DOFPose24}, is insufficient for open-world manipulation guided by language instructions~\cite{LanguageInstructRobots11,SayCan22}. We introduce semantic orientation, linking orientational vectors of an object to open-vocabulary prompts (\eg, the ``handle'' direction of a knife or ``plug-in'' direction of a USB). This bridges geometric reasoning with functional semantics, enabling robots to interpret task-specific orientation changes.
\vspace{-3pt}
\end{itemize}

\begin{figure*}[t!]
  \begin{center}
  % \vspace{-10pt}
  \includegraphics[width=1.0\linewidth]{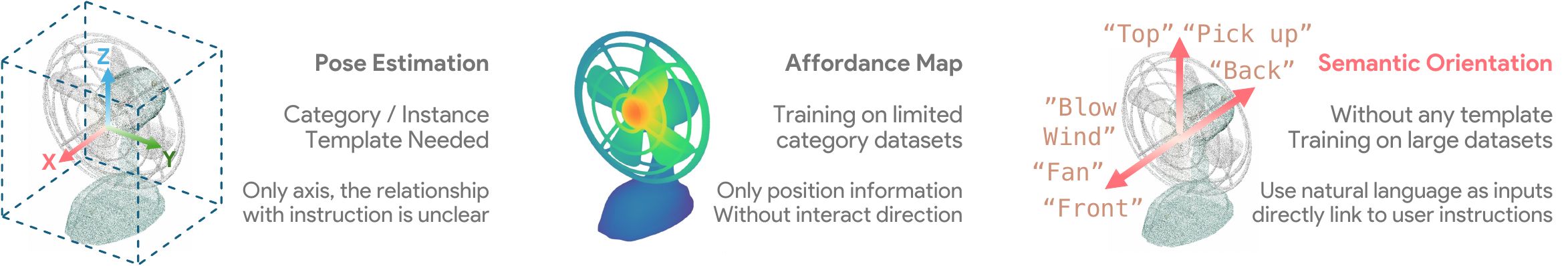}
  \vspace{-12pt}
  \caption{\textbf{Representation comparison between semantic orientation and others}.}
  \label{fig:semantic_orientation}
  \end{center}
  \vspace{-10pt}
\end{figure*}

Achieving such open-world orientation understanding requires rich world knowledge. To this end, we design both the model architecture and the dataset accordingly. We propose \textbf{PointSO}, a generalizable cross-modal 3D Transformer~\cite{Transformer17,ACT23,ReCon23,ShapeLLM24} for semantic orientation prediction. To train it at scale, we construct \textbf{OrienText300K}, a large-scale dataset comprising over 350K 3D models with diverse orientation-text pairs. These annotations are from Objaverse~\cite{objaverse23} and
generated automatically by prompting GPT-4o~\cite{GPT4o24} with rich semantic queries covering both intra-object spatial reasoning and inter-object manipulation contexts—eliminating the need for costly robot-collected data. 

To enable comprehensive spatial reasoning, we develop \textbf{\sofar}, an integrated system that combines PointSO with foundation models such as SAM~\cite{SAM23}. Given an RGB-D input, SAM segments the scene, and PointSO estimates object orientations to build an orientation-aware 3D scene graph. The graph together with the image is fed into a VLM to generate chain-of-thought~\cite{CoT22} spatial reasoning, supporting both positional and orientational planning for downstream robotic manipulation.

In addition, we introduce Open6DOR V2, a large-scale benchmark for 6-DoF object rearrangement in simulation, which supports both open-loop and closed-loop control. Our method significantly outperforms state-of-the-art VLMs and VLA models—even those trained on expensive robot trajectories—across both simulated and real-world tasks. We also introduce 6-DoF SpatialBench, a new spatial visual-question-answering benchmark to rigorously assess orientation-aware reasoning.

In summary, we propose \textbf{Semantic Orientation} as a new representation that bridges spatial reasoning and robotic manipulation, enabling open-vocabulary, template-free orientation understanding for unseen objects. We introduce \textbf{OrienText300K}, a large-scale dataset including 350K diverse objects \& orientations and 8M images through careful filtering and annotating. We develop the \textbf{\sofar} system, which enhances spatial reasoning with 6-DoF scene graph and achieves SOTA performance on Open6DOR, SimplerEnv, and generalizes across embodiments (\eg, grippers, suction cups, dexterous hands) and tasks (\eg, manipulation, navigation, VQA) without any task-specific fine-tuning. Finally, we present two new benchmarks, \textbf{Open6DOR V2} and \textbf{6-DoF SpatialBench}, to evaluate 6-DoF rearrangement and spatial reasoning.

% \begin{figure}[t!]
%   \begin{center}
%   \includegraphics[width=0.6\linewidth]{figs/src/data_construction.pdf}
%   \caption{\textbf{Data Construction of OrienText300K}.} \label{fig:data_construction}
%   \end{center}
%   \vspace{-10pt}
% \end{figure}

\begin{figure}[t!]
  \centering
  \vspace{-5pt}
  \subfloat[Data Construction of OrienText300K.]{%
  \includegraphics[width=0.48\linewidth]{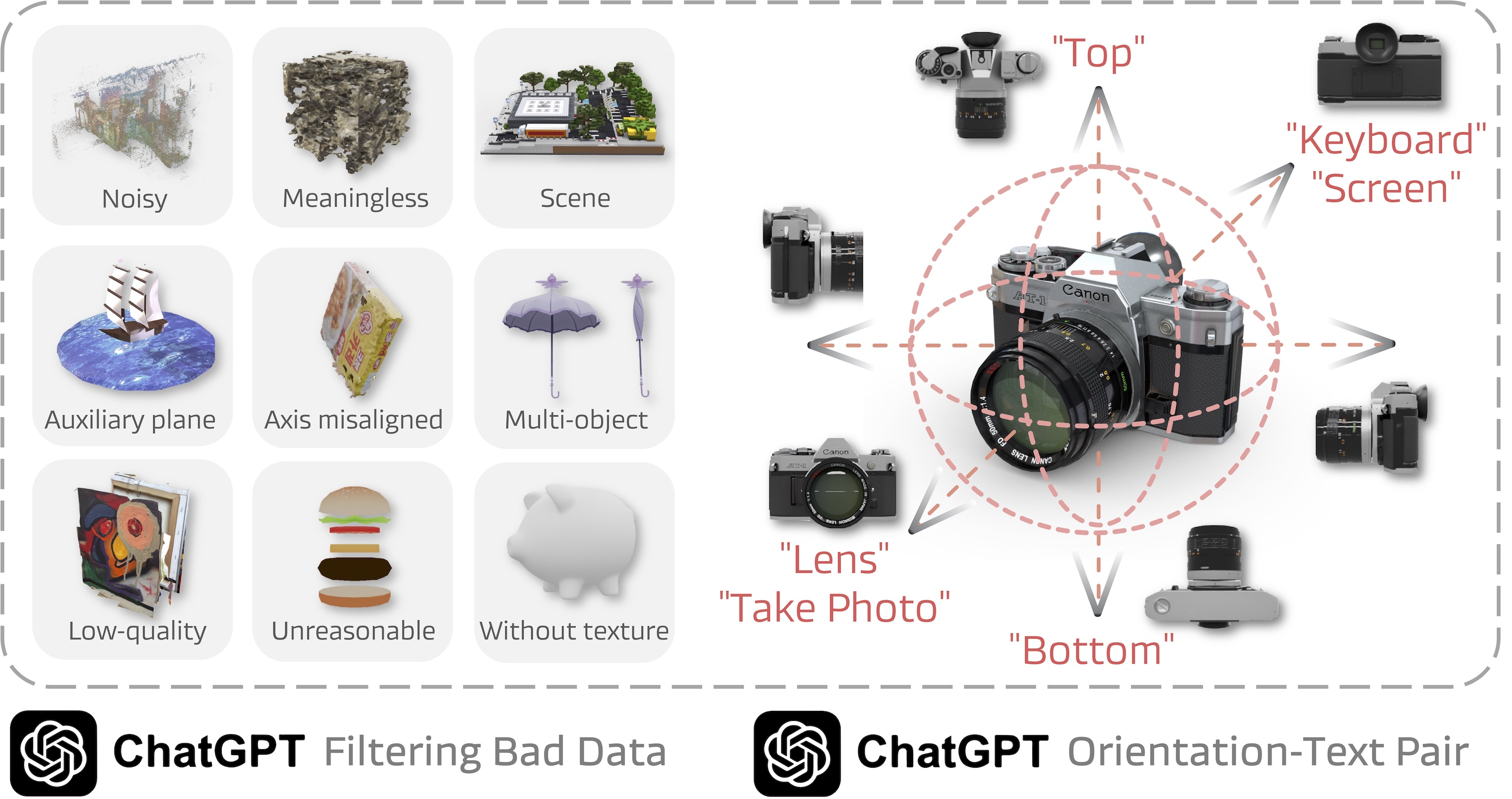}
  \label{fig:data_construction}}
  \hfill
  \subfloat[Data filtering and annotating accuracy.
  % SO denotes the annotation quality for semantic orientation. All VLMs achieve high accuracy and GPT-4o consistently yields the best result.
  ]{%
    \includegraphics[width=0.48\linewidth]{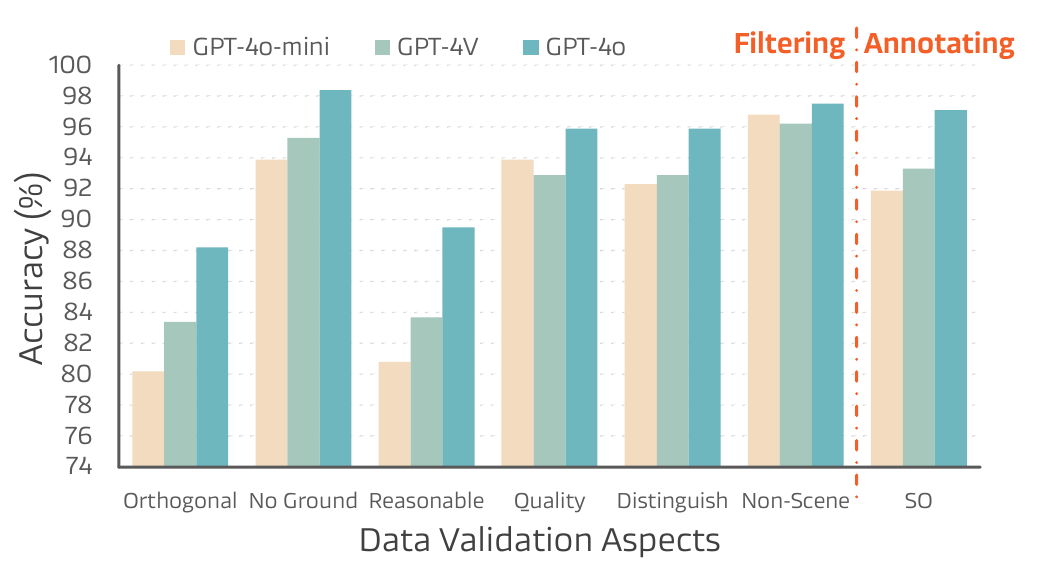}\label{fig:data_val}}
  \vspace{-2pt}
  \caption{\textbf{Visualization of OrienText300K data construction and validation results}.}
  \label{fig:data_overview}
  \vspace{-10pt}
\end{figure}

\section{Semantic Orientation: Connecting Language and Object Orientation}\label{sec:SO_def}
\subsection{Definition of Semantic Orientation}
Traditionally, object orientation is defined within a reference frame using quaternions or Euler angles to describe relative rotations. However, in interactive tasks, orientations often carry semantic meaning. Humans naturally interpret orientation in a semantic, reference-free manner. For example, plugging in a charger involves aligning the metal prongs with the socket’s opening direction—a semantically grounded alignment.
Motivated by this, we define an object’s \textit{Semantic Orientation} as a unit vector that captures the direction corresponding to a given language description. Formally, for an object $X$ and a description $\ell$, the semantic orientation $\mathbf{s}_{\ell}^X \in S(2)$ is defined as:
\begin{equation}
    \mathbf{s}_{\ell}^X = \mathcal{F}(X, \ell).
\end{equation}
Here, $\ell$ is open-vocabulary phrase referring to general directions (\eg, \textit{front}, \textit{top}), object parts (\eg, \textit{handle}, \textit{cap}), or interactions (\eg, \textit{pour out}, \textit{plug-in}).
An object $X$ can be associated with multiple semantic orientations by varying the language input, forming a set $S_X = \{ \mathbf{s}_{\ell_1}^X, \mathbf{s}_{\ell_2}^X, \dots, \mathbf{s}_{\ell_n}^X \}$. These orientations provide a semantic basis for describing and transforming the object's rotation.

\subsection{OrienText300K: Orientation-Text Paired Data at Scale}
Our goal is to develop an \textit{orientation model} capable of identifying semantic orientations in open-world settings using large-scale 3D data. To support this, we introduce OrienText300K, a curated dataset of 3D models annotated with diverse language-guided orientation labels. The dataset is constructed from Objaverse~\cite{objaverse23}, which contains approximately 800K Internet-sourced 3D models across a wide range of categories. Since the raw data includes noisy annotations and low-quality samples, we apply a rigorous filtering process. Using Blender, we render over 8M high-quality images under carefully designed lighting conditions to ensure fidelity for training.

\noindent{\textbf{Data Filtering}}~
To ensure high-quality data for generating semantic orientation annotations, we apply a dedicated filtering strategy that retains only the samples meeting the following six criteria.
\ding{182} Standard orthogonal view only. Samples in random views will be filtered.
\ding{183} Clean objects without the ground for auxiliary visualization.
\ding{184} Reasonable objects that have sufficient spatial reasoning potentials.
\ding{185} High-quality objects. Blurry and wrong samples are filtered.
\ding{186} Distinguishable objects. Abstract and meaningless objects are filtered.
\ding{187} Non-scene objects for object-centric understanding.

However, it is non-trivial to conduct filtering on such big data using manual labor.
Inspired by recent works showing large VLMs are human-aligned judgers~\cite{LLMAsJudge23,GPT4V-3DEvaluator24,DreamBenchPlus24}, we employ GPT-4o~\cite{GPT4o24} by prompting requirements above.
To be specific, the multi-view images of 3D objects are concatenated together with our designed prompts into GPT-4o, and GPT-4o will decide whether samples should be filtered.
The filtered dataset yields 350K+ clean samples, significantly reducing data noise.

\noindent{\textbf{Data Annotation}}~
As mentioned in the introduction, VLMs struggle to produce accurate object orientation values, which presents a significant challenge for data generation. Fortunately, VLMs are powerful discriminators capable of distinguishing between different views through multimodal understanding. We believe that the initial stage of data cleaning effectively removed a large amount of misaligned data, leaving behind a set of properly aligned instances capable of producing \textit{standard} orthogonal views. We then leverage GPT-4o to interpret the semantic content across six views and generate semantic-view pairs accordingly. 
Throughout the annotation process, both human modelers in Objaverse and ChatGPT serve as our annotators, supplying the necessary knowledge to produce both view-aligned data and semantically grounded annotations.

\noindent{\textbf{Quality Validation}}~
To validate annotation quality, we construct a validation set containing 208 samples with manually labeled filtering criteria and semantic orientation labels, respectively.
From \cref{fig:data_val}, we observe that GPT-4o achieves an average accuracy of 88.3\% and 97.1\% accuracy on filtering and annotating, respectively.
This provides a quality guarantee of our OrienText300K.

\subsection{PointSO: A Cross-Modal 3D Transformer for Semantic Orientation Prediction}\label{sec:PointSO}
\begin{wrapfigure}{r}{0.5\linewidth}
  \centering
  \vspace{-12pt}
  \includegraphics[width=\linewidth]{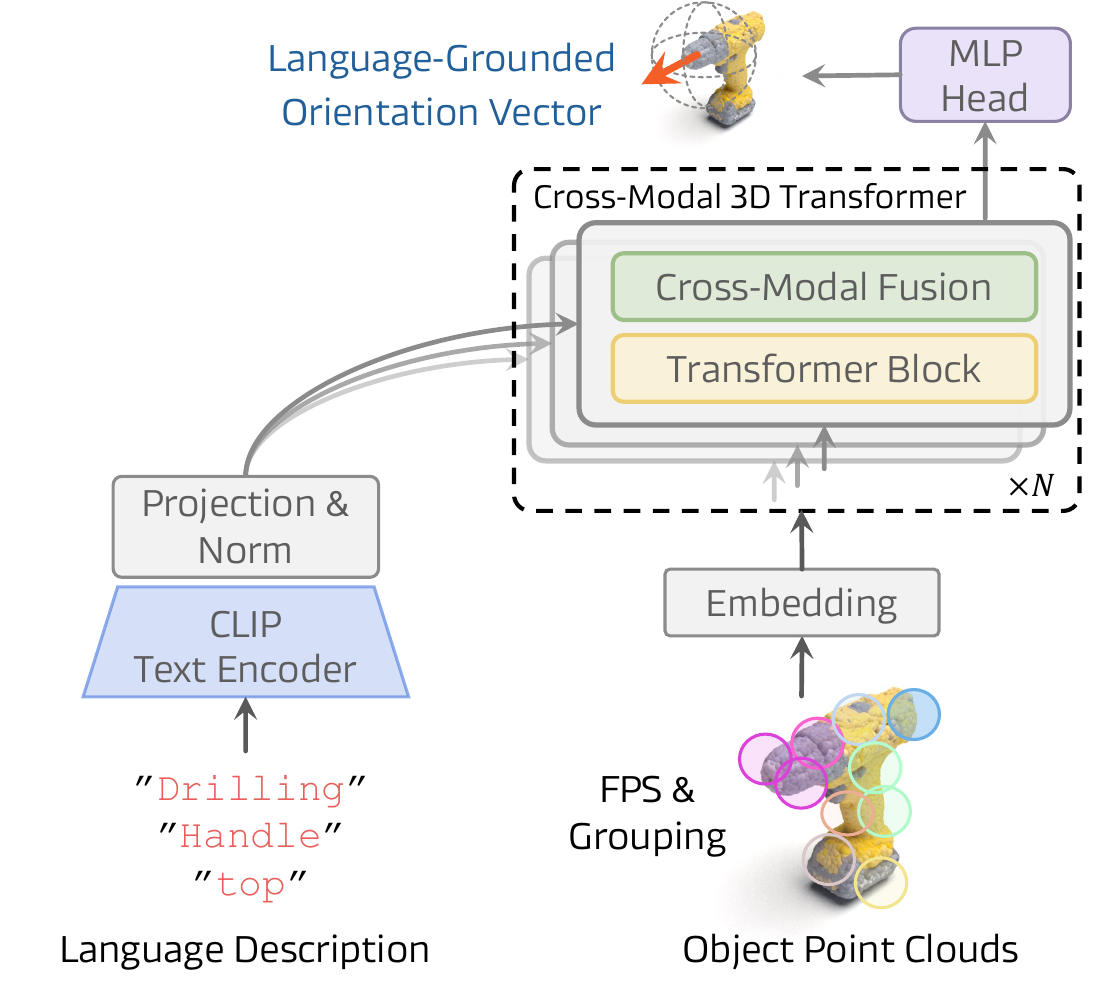}
  \vspace{-15pt}
  \caption{\textbf{PointSO model architecture}.}
  \vspace{-10pt}
  \label{fig:PointSO}
\end{wrapfigure}

We introduce PointSO, a plain Transformer-based architecture~\cite{Transformer17} with cross-modal 3D-language fusion as our orientation model. 
As illustrated in \cref{fig:PointSO}, PointSO takes the object's 3D point clouds and a language description as inputs, and predicts the corresponding semantic orientation.

% \begin{figure}[t!]
%   \begin{center}
%   \includegraphics[width=0.6\linewidth]{figs/src/pointso.pdf}
%   \vspace{-10pt}
%   \caption{\textbf{PointSO architecture} for semantic orientation prediction.
%   % PointSO takes 3D and language embeddings as inputs of a $N$-layer plain Transformer model, and uses a token-level sum as cross-modal fusion in all layers.
%   % An MLP head predicts the corresponding orientation vector normalized in a unit sphere.
%   }
%   \label{fig:PointSO}
%   \end{center}
%   \vspace{-12pt}
% \end{figure}

\noindent{\textbf{3D and Language Embeddings}}~
Given an object's point cloud $X = \{\mathbf{x}_i \in\mathbb{R}^3|i=1,2,\dots, N\}$ with $N$ 3D points defined in (x, y, z) Cartesian space, and an arbitrary language description $\ell$, we first embed both into discrete token embeddings.
For the 3D point clouds, we follow~\cite{ACT23,PointBERT22,ReCon23} to first sample $N_s$ seed points using farthest point sampling (FPS) and then group inputs with KNN for point feature embedding with a local geometric extraction network such as lightweight PointNet~\cite{PointNet17,PointNet++17}. 
An MLP head is used which maps a special \texttt{[CLS]} token~\cite{ViT21} to a predicted direction.
As for the language inputs, we adopt CLIP~\cite{CLIP21} and use the global token as cross-modal fusion inputs.

\noindent{\textbf{Cross-Modal Fusion}}~
We perform cross-modal fusion by injecting global text features into each layer of the 3D Transformer using a simple yet effective strategy: adding the text token to every point token. While other fusion methods such as cross-attention, adapters, or concatenation along spatial or channel dimensions are possible, we empirically find that token-wise addition performs best (see~\cref{app:fusion}). This effectiveness may stem from the short language inputs, where summation helps reinforce their influence across layers.

\noindent{\textbf{Optimization}}~
Let $\mathcal{F}_{\text{SO}}$ represent the PointSO model parameterized by $\theta_{\text{SO}}$ (the CLIP is kept frozen and thus its parameters are not included).
Given every object point cloud $X_i \in \mathcal{D}_{\text{OrienText300K}}$ in the OrienText300K dataset, where each object is labeled with a language set $L_i=\{\ell_j^i, j=1,2,\dots,Q\}$ and the corresponding ground truth semantic orientation set, $S_i=\{\mathbf{s}^i_j, j=1,2,\dots,Q\}$.
The optimization is to minimize the negative cosine similarity $\mathcal{L}_{\text{cos}}(\mathbf{v},\mathbf{k})=\mathbf{1}-\frac{\mathbf{v}\cdot\mathbf{k}}{\|\mathbf{v}\|\cdot\|\mathbf{k}\|}$ between predicted and the ground truth semantic orientations:
\begin{equation}
    \min_{\theta_{\text{SO}}} \sum_{X_i \in \mathcal{D}_{\text{OrienText300K}}} \sum_{\ell_j^i\in L_i} 
    \mathcal{L}_{\text{cos}}
    \Big(
    \mathcal{F}_{\text{SO}}(X_i, \ell_j^i), \mathbf{s}^i_j
    \Big).
\end{equation}
\section{\sofar: Semantic Orientation Bridges
Spatial Reasoning and Object Manipulation}
\label{sec:sofar_graph}
\begin{figure*}[t!]
  \begin{center}
  % \vspace{-15pt}
  \includegraphics[width=\linewidth]{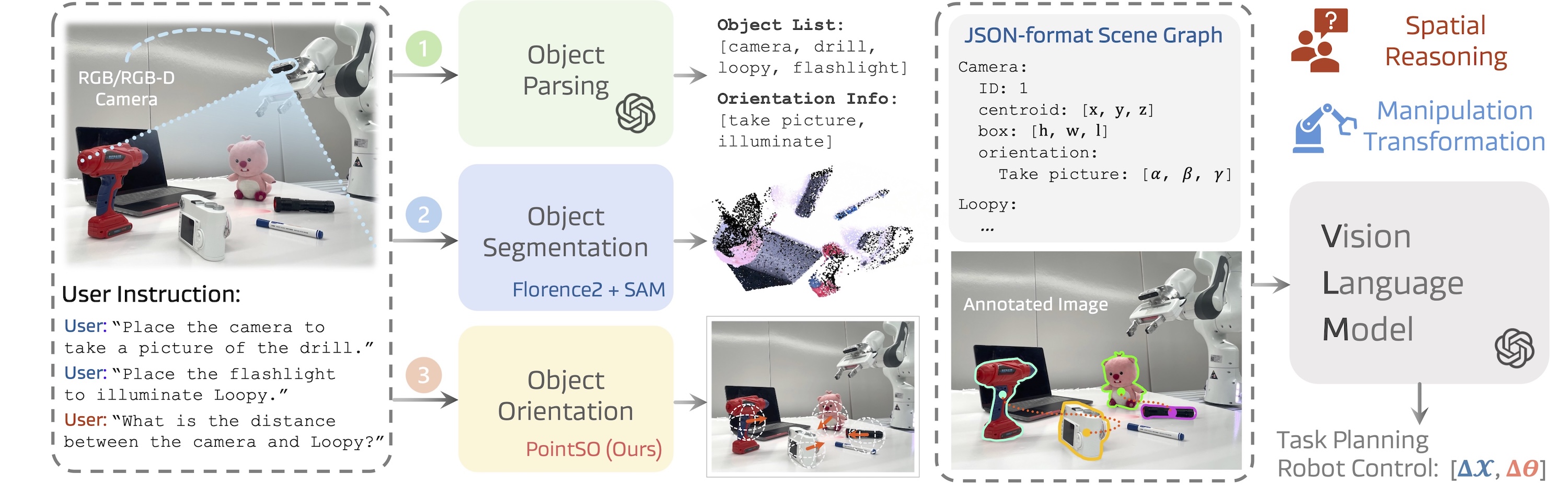}
  \vspace{-16pt}
  \caption{\textbf{Overview of \sofar~system.} Given RGB-D images and language instructions, \sofar~first leverages a VLM to identify relevant object phrases and semantic orientations. Then utilizes foundation models Florence-2~\cite{florence2}, SAM~\cite{SAM23}, and our PointSO for object segmentation and semantic orientation estimation. This information forms a 6-DoF scene graph, which the VLM uses alongside the RGB image to perform spatial understanding tasks or generate manipulation actions.
  }
  % \caption{\textbf{Overview of \sofar~system}. \sofar~takes RGB-D images as inputs, where the depth images can be obtained from depth senor or metric 3D prediction~\cite{Metric3D23}. Given the language instruction, \sofar~prompts the VLM to obtain task-oriented object phases and semantic orientation descriptions. Then, \sofar~leverage foundation models Florence-2~\cite{florence2} and SAM~\cite{SAM23} to segment depth point clouds and our PointSO (\cref{sec:PointSO}) to obtain semantic orientations. Summarizing 3D object-centric information, an orientation-aware scene graph is constructed and encoded into languages (\cref{sec:sofar_graph}). The VLM takes the RGB image and the scene graph and outputs the queried spatial understanding VQA or translation for manipulation.}
  \label{fig:pipeline}
  \end{center}
  \vspace{-5pt}
\end{figure*}

Our proposed PointSO model now paves the way for off-the-shelf object-centric spatial orientation understanding. However, it remains challenging to extend such object-centric spatial understanding for scene-level spatial reasoning both in the digital world (\eg, 6-DoF visual question answering) and in the physical world (\eg, robot manipulations). To bridge this gap, we build an integrated reasoning system where a powerful VLM acts as an agent and reasons about the scene while communicating with off-the-shelf models including PointSO and SAM~\cite{SAM23}. \cref{fig:pipeline} illustrates an overview of our proposed framework, aiming at \textbf{S}emantic \textbf{O}rientation \textbf{F}or \textbf{A}utonomous \textbf{R}obots (\textbf{\sofar}).

\subsection{Scene Graph with 6-DoF Information}
\label{sec:scene_graph}
To integrate both the positional \& orientational interaction relationships of objects, we use a scene graph with 6-DoF information to represent the environment.

\noindent{\textbf{Position \& Orientation Information Extraction}}~
Given a language query $\mathcal{Q}$, we first prompt a vision-language model $\mathcal{F}_{\text{VLM}}$ to extract a task-relevant set of object phrases $\mathcal{P} = \{p_i\,|\,i = 1, 2, \dots, M\}$. 
Each phrase $p_i$ represents a language description of an object relevant to $\mathcal{Q}$. 
Using the SAM~\cite{SAM23} \& Florence-2~\cite{florence2}, we perform language-conditioned segmentation to obtain a corresponding object set $\mathcal{X} = \{X_i\,|\,i = 1, 2, \dots, M\}$, where $X_i$ is the 3D point cloud of the $i$-th object.
Each object is assigned a unique ID for use in Set-of-Mark (SoM) prompting~\cite{SoM23}.
We then prompt the VLM to generate a set of task-specific orientation descriptions $L_i$ for related objects, and use pretrained PointSO to infer their semantic orientations, resulting in a semantic orientation set $S_i$.

\noindent{\textbf{6-DoF Scene Graph}}~
From the segmented object set $\mathcal{X}$, we construct an 6-DoF scene graph $\mathcal{G} = (\mathbf{V}, \mathbf{E})$ with $M$ nodes.
Each node $\mathbf{o}_i \in \mathbf{V}$ encodes the following semantic and spatial attributes:
\ding{182} object phrase $p_i$ with a unique instance ID;
\ding{183} 3D position $\mathbf{c}_i = (x, y, z) \in \mathbb{R}^3$ from the object's centroid;
\ding{184} bounding box size $\mathbf{b}_i = (h, w, l) \in \mathbb{R}^3$;
\ding{185} semantic orientation set $S_i$ along with its corresponding description set $L_i$.
Each edge $\mathbf{e}_{ij} \in \mathbf{E}$ represents the relative translation and size ratio between two connected objects $\mathbf{o}_i$ and $\mathbf{o}_j$.

\subsection{Spatial-Aware Task Reasoning}
\label{sec:manip_pipline}

We encode the 6-DoF scene graph $\mathcal{G}$ into descriptive language and input it to the VLM alongside the RGB image $I$ and query $\mathcal{Q}$. This enriched spatial representation enables the VLM to perform accurate spatial reasoning by leveraging its visual and linguistic understanding.

\noindent{\textbf{Chain-of-Thought Spatial Reasoning}}~
Most robot manipulation tasks involving rigid objects can be abstracted as applying transformations to adjust their position and orientation. To guide the VLM in generating such transformations from language instructions, we adopt a CoT reasoning process~\cite{CoT22} that decomposes the reasoning into three steps: 
(i) analyzing the scene with the query $\mathcal{Q}$ and object nodes $\mathbf{V}$;
(ii) computing the desired position and orientation of the target object;
(iii) predicting the target position $\Tilde{\mathbf{c}}_i$ and semantic orientation set $\Tilde{S}_i$ for each object.
Given the initial state $\mathbf{c}_i$ and $S_i$, the full 6-DoF transformation $\mathbf{P}_i$ is computed. Specifically, translation is obtained by $\mathbf{t}_i = \Tilde{\mathbf{c}}_i - \mathbf{c}_i$, and rotation $\mathbf{R}_i$ is estimated from $S_i$ and $\Tilde{S}_i$ using the Kabsch-Umeyama algorithm~\cite{Kabsch76,Kabsch78,least91}.

\noindent{\textbf{Low-Level Motion Execution}}~
\label{execution}
Following CoPa~\cite{CoPa24}, we integrate task-specific grasping and motion planning. Object or part segmentation is performed using Florence-2~\cite{florence2} and SAM~\cite{SAM23}, followed by grasp candidate generation via GSNet~\cite{GraspNet1B20}. The optimal grasp is selected by considering both grasp quality and heuristics. Based on instruction, \ours~predicts the object's translation and rotation, defining the transformation from grasp to placement. We employ OMPL~\cite{ompl} to generate a collision-free trajectory, initializing joint positions at the midpoint to ensure smooth and safe motion.

\begin{figure*}[t!]
\begin{center}
\vspace{-4pt}
\includegraphics[width=\linewidth]{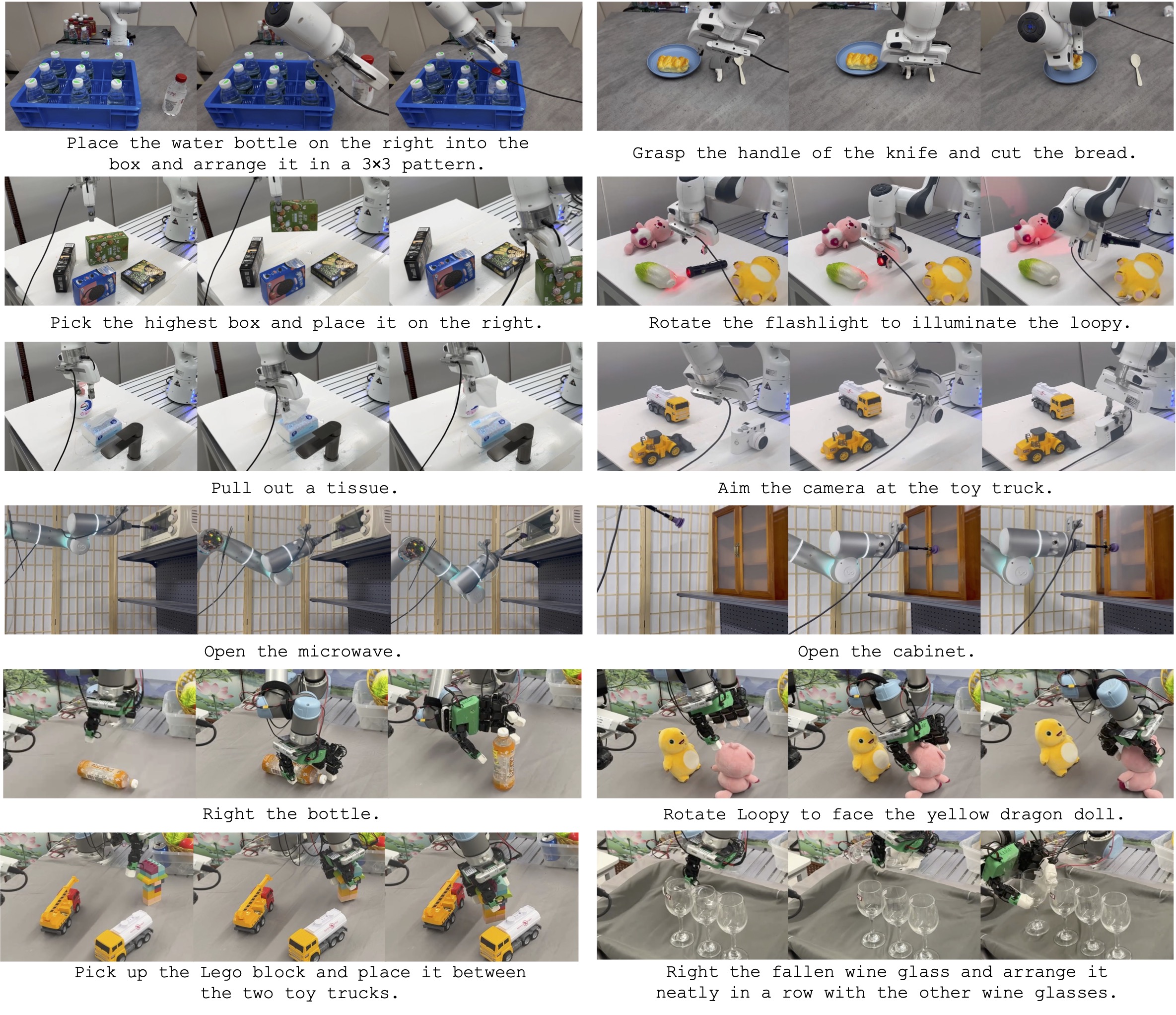}
\vspace{-16pt}
\caption{\textbf{Qualitative results} of real world language-grounded manipulation. \sofar~can generalize across various \textbf{embodiments, tasks and environments}.
}
\vspace{-12pt}
\label{fig:real_demo}
\end{center}
\end{figure*}

\begin{figure*}[t!]
  \begin{center}
  \includegraphics[width=1.0\linewidth]{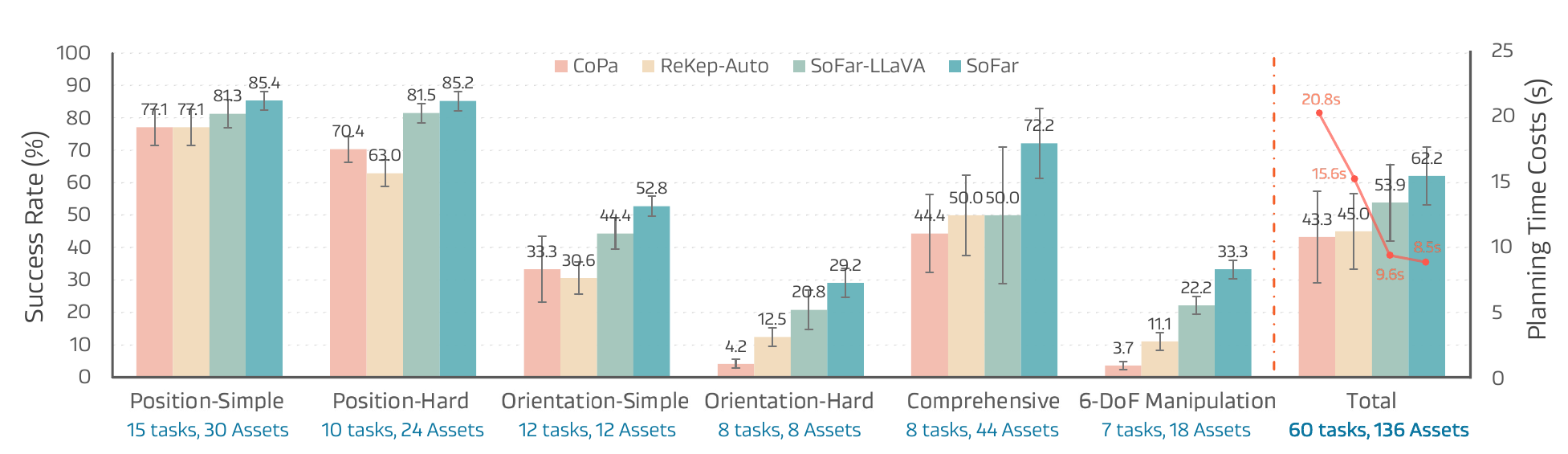}
  \vspace{-15pt}
  \caption{\textbf{Quantitative evaluation} of zero-shot real-world language-grounded rearrangement. We design \textbf{60} diverse real-world tasks involving over \textbf{100} diverse objects (detailed in \cref{tab:detailed_realworld}). 
  } \label{fig:real}
  \end{center}
  \vspace{-10pt}
\end{figure*}

\section{Experiments}
\label{sec:exp}

\subsection{Real-world Language-Grounded Object Manipulation}

\noindent{\textbf{Tasks and Evaluations}}~
We construct 60 real-world tasks involving over 100 objects, following the Open6DOR benchmark~\cite{Open6DOR24}. The tasks are divided into three tracks—position, orientation, and comprehensive \& 6-DoF—each with simple and hard variants. The position track assesses spatial reasoning from basic (\eg, front/back/left/right) to more complex relations (\eg, between/center/custom). The orientation track includes part-level orientation in the simple setting, and fine-grained angle estimation in the hard setting. The comprehensive and 6-DoF tracks evaluate complex instruction understanding and simultaneous control over position and orientation. Each task is repeated three times to ensure statistical robustness. More details and visualizations are available in~\cref{app:detail_realworld}.

\noindent{\textbf{Results}}~
As shown in \cref{fig:real}, \sofar~consistently outperforms baselines across all tracks, especially on orientation and 6-DoF tasks, while maintaining low planning overhead. We also demonstrate \sofar's embodiment generality with different end-effectors, including dexterous hands and suction cups, as illustrated in \cref{fig:real_demo}. Additional robot setups and generalization results are provided in~\cref{app:robot_setup}.

\begin{table*}[t!]
\setlength{\tabcolsep}{3.5pt}
% \vspace{-12pt}
\caption{\textbf{6-DoF object rearrangement evaluation} on Open6DOR~\cite{Open6DOR24}.}
\vspace{-4pt}
\label{tab:open6dor}
\centering
\resizebox{\linewidth}{!}{
\begin{tabular}{lccccccccccc}
\toprule[0.95pt]
\multirow{2}{*}[-0.5ex]{Method} & \multicolumn{3}{c}{\textbf{Position Track}} & \multicolumn{4}{c}{\textbf{Rotation Track}} & \multicolumn{3}{c}{\textbf{6-DoF Track}} & \multirow{2}{*}[-0.5ex]{Time Cost (s)}\\
\cmidrule(lr){2-4} \cmidrule(lr){5-8} \cmidrule(lr){9-11}
& Level 0 & Level 1 & Overall & Level 0 & Level 1 & Level 2 & Overall & Position & Rotation & Overall \\ 
\midrule[0.6pt]
\multicolumn{12}{c}{\textit{Perception Tasks on Issac Sim~\cite{IsaacSim21} (Open6DOR V1 Setting)}}\\
\midrule[0.6pt]
GPT-4V~\cite{GPT4Vision23} & 46.8 & 39.1 & 45.2 & 9.1 & 6.9 & 11.7 & 9.2 & - & - & - & - \\
Dream2Real~\cite{Dream2Real24} & 17.2 & 11.0 & 15.9 & 37.3 & 27.6 & 26.2 & 31.3 & 26.2 & 18.7 & 13.5 & 358.3s \\
VoxPoser~\cite{VoxPoser23} & 35.6 & 21.7 & 32.6 & - & - & - & - & - & - & -  & -\\
Open6DOR-GPT~\cite{Open6DOR24} & 78.6 & 60.3 & 74.9 & 45.7 & 32.5 & 49.8 & 41.1 & 84.8 & 40.0 & 35.6 & 126.3 s\\
\rowcolor{linecolor1}\textbf{\ours-LLaVA} & 86.3 & 57.9 & 78.7 & 62.5 & 30.2 & 67.1 & 48.6 & 83.0 & 48.2 & 40.3 & 9.6s\\
\rowcolor{linecolor2}\textbf{\ours} & \textbf{96.0} & \textbf{81.5} & \textbf{93.0} & \textbf{68.6} & \textbf{42.2} & \textbf{70.1} & \textbf{57.0} & \textbf{92.7} & \textbf{52.7} & \textbf{48.7} & \textbf{8.5s}\\
\midrule[0.6pt]
\multicolumn{12}{c}{\textit{Execution Tasks on Libero~\cite{LIBERO23} (Open6DOR V2 Setting)}}\\
\midrule[0.6pt]
Octo~\cite{Octo24} & 51.2 & 32.1 & 47.2 & 10.7 & 18.3 & 29.9 & 17.2 & 45.6 & 8.0 & 8.0 & - \\
OpenVLA~\cite{OpenVLA24} & 51.6 & 32.4 & 47.6 & 11.0 & 18.5 & 30.6 & 17.6 & 46.2 & 8.2 & 8.2 & - \\
\rowcolor{linecolor2}\textbf{\ours} & \textbf{72.1} & \textbf{47.6} & \textbf{67.0} & \textbf{28.3} & \textbf{18.3} & \textbf{34.7} & \textbf{25.7} & \textbf{63.7} & \textbf{25.6} & \textbf{18.4} & \textbf{40s}\\
\bottomrule[0.95pt]
\end{tabular}
}
\end{table*}

\begin{wraptable}{r}{0.5\textwidth}
\vspace{-14pt}
\setlength{\tabcolsep}{7.5pt}
\caption{\textbf{Semantic Orientation evaluation} on OrienText300K validation split.}
\label{tab:semantic_orientation}
\centering
\vspace{-4pt}
\resizebox{0.48\textwidth}{!}{
\begin{tabular}{lccccc}
\toprule[0.95pt]
    Method & \texttt{45°} & \texttt{30°} & \texttt{15°} & \texttt{5°} & Avg. \\ 
    \midrule[0.6pt]
    PointSO-S & 77.34 & 74.22 & 67.97 & 60.94 & 70.12 \\
    \rowcolor{linecolor1}\textbf{PointSO-B} & \textbf{79.69} & \textbf{77.34} & \textbf{70.31} & \textbf{62.50} & \textbf{72.46} \\
    \rowcolor{linecolor2}\textbf{PointSO-L} & \textbf{81.25} & \textbf{78.13} & \textbf{72.66} & \textbf{65.63} & \textbf{74.42} \\
\bottomrule[0.95pt]
\end{tabular}
}
\vspace{9pt}
\setlength{\tabcolsep}{6pt}
\caption{\textbf{Semantic Orientation evaluation} of \textit{robustness}. 
\texttt{Single-View}: randomly select a camera viewpoint within the unit sphere and generate a single FoV viewpoint in polar coordinates.
\texttt{Jitter}: Gaussian noise $\epsilon\sim\mathcal{N}(0,\sigma^2)$, $\sigma=0.01$.
\texttt{Rotate}: random SO(3) rotation $(\alpha,\beta,\gamma)\sim \mathcal{U}(-\pi,\pi)$.
\texttt{All}: all corruptions.
}
\vspace{-3pt}
\label{tab:semantic_orientation_c}
\centering
\resizebox{0.48\textwidth}{!}{
\begin{tabular}{lcccc}
\toprule[0.95pt]
\multirow{2}{*}[-0.5ex]{Method} & \multicolumn{4}{c}{\textbf{OrienText300K-C Variants}} \\
\cmidrule(lr){2-5}
& \texttt{Single}-\texttt{View} & \texttt{Jitter} & \texttt{Rotate} & \texttt{All} \\
\midrule[0.6pt]
PointSO-S & 72.66 & 76.56 & 73.43 & 67.19 \\
\rowcolor{linecolor1}\textbf{PointSO-B} & \textbf{75.00} & \textbf{78.90} & \textbf{75.78} & \textbf{71.09} \\
\rowcolor{linecolor2}\textbf{PointSO-L} & \textbf{76.56} & \textbf{81.25} & \textbf{77.34} & \textbf{74.22} \\
\bottomrule[0.95pt]
\end{tabular}
}
\vspace{-15pt}
\end{wraptable}

\subsection{Semantic Orientation Prediction}
Using free-text descriptions to extract semantic orientations from object point clouds is challenging. In Objaverse~\cite{objaverse23}, we manually annotate 128 diverse objects and construct the OrienText300K val split to evaluate the directional prediction accuracy of PointSO. We train different model variants on OrienText300K, and the results in \cref{tab:semantic_orientation} report performance across different angular thresholds ranging from 45° to 5°. PointSO still has an accuracy rate of 60\% even under a 5° threshold.

In the real world, obtaining complete object point clouds is often difficult. To evaluate the robustness of PointSO under such conditions, we introduce three types of input perturbations: random rotations, partial single-sided observations, and Gaussian noise. As reported in \cref{tab:semantic_orientation_c}, the accuracy at the 45° threshold reflects the model's resilience to these corruptions.

\subsection{6-DoF Object Rearrangement Evaluation on Open6DOR V2}
To evaluate 6-DoF object rearrangement capabilities, we extend the original Open6DOR benchmark~\cite{Open6DOR24}, which primarily focuses on final pose estimation, into a more comprehensive setting that includes both perception and execution evaluation. We migrate its scenes into a robosuite-based simulation environment~\cite{robosuite2020}, following the task interface defined by LIBERO~\cite{LIBERO23}, and name this new benchmark Open6DOR V2.
Results are reported in~\cref{tab:open6dor}. For perception tasks, we adopt the original Open6DOR~\cite{Open6DOR24} evaluation protocol and compare with the same baselines. \ours~achieves the best performance, demonstrating strong spatial understanding and zero-shot generalization. For execution tasks, we compare against the pretrained Octo~\cite{Octo24} and the LIBERO-finetuned OpenVLA~\cite{OpenVLA24}, all evaluated in the same robosuite environment to minimize domain shift. While both baselines show limited success due to poor generalizability, \ours~reaches around 40\% success rate using a vanilla execution pipeline. We note that certain objects are intrinsically difficult to manipulate, suggesting the need for more robust policies incorporating prehensile grasping and adaptive strategies to improve performance on Open6DOR V2.

\begin{table*}[t!]
\centering
\setlength{\tabcolsep}{1.2pt}
\caption{\textbf{SimplerEnv~\cite{simplerenv24} simulation evaluation results for the Google Robot setup.}
We present success rates for the ``Variant Aggregation'' and ``Visual Matching'' approaches.
Top-1 \& Top-2 accuracies are represented using different colors.
OXE: Open X-Embodiment dataset~\cite{OpenXEmbodiment24}.
}
\vspace{-2pt}
\scriptsize
\resizebox{1.0\linewidth}{!}{
\begin{tabular}{llcccccccccc}
\toprule
\multirow{5}{*}{\begin{tabular}[l]{@{}l@{}}Google Robot\\ Evaluation Setup\end{tabular}} &
\multirow{5}{*}{Policy} &
\multirow{5}{*}{Training Data} &
\multicolumn{4}{c}{\multirow{2}{*}{Pick Coke Can}} &
\multirow{2}{*}{Move Near} &
\multicolumn{3}{c}{\multirow{2}{*}{Open / Close Drawer}}  & \multirow{5}{*}{Average}\\ \\ \cmidrule{4-11}
& & & 
\begin{tabular}[c]{@{}c@{}}Horizontal\\ Laying\end{tabular} &
\begin{tabular}[c]{@{}c@{}}Vertical\\ Laying\end{tabular} &
\begin{tabular}[c]{@{}c@{}}Standing\end{tabular} &
\begin{tabular}[c]{@{}c@{}}Average\end{tabular} &
\begin{tabular}[c]{@{}c@{}}Average\end{tabular} &
\begin{tabular}[c]{@{}c@{}}Open\end{tabular} &
\begin{tabular}[c]{@{}c@{}}Close\end{tabular} &
\begin{tabular}[c]{@{}c@{}}Average\end{tabular} \\
\midrule
\multirow{5}{*}[-0.5ex]{\begin{tabular}[l]{@{}l@{}} Variant\\Aggregation \end{tabular}} 
& RT-1-X~\cite{OpenXEmbodiment24} & OXE & 0.569 & 0.204 & 0.698 & 0.490 & 0.323 & 0.069 & \cellcolor{linecolor2}{\textbf{0.519}} & 0.294 & 0.397\\
& RT-2-X~\cite{RT223} & OXE & \cellcolor{linecolor1}{\underline{0.822}} & \cellcolor{linecolor1}{\underline{0.754}} & \cellcolor{linecolor1}{\underline{0.893}} & \cellcolor{linecolor1}{\underline{0.823}} & \cellcolor{linecolor2}{\textbf{0.792}} & \cellcolor{linecolor2}{\textbf{0.333}} & 0.372 & \cellcolor{linecolor2}{\textbf{0.353}} & \cellcolor{linecolor1}{\underline{0.661}}\\ 
& Octo-Base~\cite{Octo24} & OXE & 0.005 & 0.000 & 0.013 & 0.006 & 0.031 & 0.000 & 0.021 & 0.011 & 0.012\\ 
& OpenVLA~\cite{OpenVLA24} & OXE & 0.711 & 0.271 & 0.653 & 0.545 & 0.477 & 0.158 & 0.195 & 0.177 & 0.411\\
\cmidrule{2-12}
& \textbf{\ours} & \textbf{Zero-Shot} & \cellcolor{linecolor2}{\textbf{0.861}} & \cellcolor{linecolor2}{\textbf{0.960}} & \cellcolor{linecolor2}{\textbf{0.901}} & \cellcolor{linecolor2}{\textbf{0.907}} & \cellcolor{linecolor1}{\underline{0.740}} & \cellcolor{linecolor1}{\underline{0.200}} & \cellcolor{linecolor1}{\underline{0.394}} & \cellcolor{linecolor1}{\underline{0.297}} & \cellcolor{linecolor2}{\textbf{0.676}} \\
\midrule
\multirow{5}{*}[-0.5ex]{\begin{tabular}[l]{@{}l@{}} Visual\\Matching \end{tabular}} 
& RT-1-X~\cite{OpenXEmbodiment24} & OXE & \cellcolor{linecolor2}{\textbf{0.820}} & 0.330 & 0.550 & 0.567 & 0.317 & \cellcolor{linecolor2}{\textbf{0.296}} & \cellcolor{linecolor2}{\textbf{0.891}} & \cellcolor{linecolor2}{\textbf{0.597}} & 0.534\\ 
& RT-2-X~\cite{RT223} & OXE & 0.740 & \cellcolor{linecolor1}{\underline{0.740}} & \cellcolor{linecolor1}{\underline{0.880}} & \cellcolor{linecolor1}{\underline{0.787}} & \cellcolor{linecolor1}{\underline{0.779}} & 0.157 & 0.343 & 0.250 & \cellcolor{linecolor1}{\underline{0.606}}\\ 
& Octo-Base~\cite{Octo24} & OXE & 0.210 & 0.210 & 0.090 & 0.170 & 0.042 & 0.009 & 0.444 & 0.227 & 0.168\\ 
& OpenVLA~\cite{OpenVLA24} & OXE & 0.270 & 0.030 & 0.190 & 0.163 & 0.462 & 0.194 & 0.518 & 0.356 & 0.277 \\
\cmidrule{2-12}
& \textbf{\ours} & \textbf{Zero-Shot} & \cellcolor{linecolor1}{\underline{0.770}} & \cellcolor{linecolor2}{\textbf{1.000}} & \cellcolor{linecolor2}{\textbf{1.000}} & \cellcolor{linecolor2}{\textbf{0.923}} & \cellcolor{linecolor2}{\textbf{0.917}} & \cellcolor{linecolor1}{\underline{0.227}} & \cellcolor{linecolor1}{\underline{0.578}} & \cellcolor{linecolor1}{\underline{0.403}} & \cellcolor{linecolor2}{\textbf{0.749}}\\
\bottomrule
\end{tabular}
\vspace{-4pt}
}
\label{tab:simpler_env}
\end{table*}

\begin{table*}[t!]
\scriptsize
\centering
\caption{\textbf{SimplerEnv~\cite{simplerenv24} simulation evaluation results for the WidowX + Bridge setup.}
We report both the final success rate (``Success'') along with partial success (\eg, ``Grasp Spoon'').
% Top-1 \& Top-2 accuracies are represented using different colors, bold text, and underlines.
OXE: Open X-Embodiment dataset~\cite{OpenXEmbodiment24}. Bridge: BridgeData V2 dataset~\cite{bridgedata23} (In domain training).}
\vspace{-2pt}
\setlength{\tabcolsep}{3.2pt}
\resizebox{1.0\linewidth}{!}{
\begin{tabular}{lcccccccccc}
\toprule
 \multirow{5}{*}{Policy} & \multirow{5}{*}{Training Data} &
 \multicolumn{2}{c}{Put Spoon} &
 \multicolumn{2}{c}{Put Carrot} &
 \multicolumn{2}{c}{Stack Green Block} & \multicolumn{2}{c}{Put Eggplant} & \multirow{5}{*}{Average}\\ 
 & &
 \multicolumn{2}{c}{on Towel} &
 \multicolumn{2}{c}{on Plate} &
 \multicolumn{2}{c}{on Yellow Block} & \multicolumn{2}{c}{in Yellow Basket}\\ \cmidrule{3-10}
 & &
 \begin{tabular}[c]{@{}c@{}}Grasp \\ Spoon\end{tabular} &
 \begin{tabular}[c]{@{}c@{}} Success\end{tabular} &
 \begin{tabular}[c]{@{}c@{}}Grasp \\ Carrot\end{tabular} &
 \begin{tabular}[c]{@{}c@{}} Success\end{tabular} &
 \begin{tabular}[c]{@{}c@{}}Grasp \\ Green Block\end{tabular} &
 \begin{tabular}[c]{@{}c@{}} Success\end{tabular} & 
 \begin{tabular}[c]{@{}c@{}}Grasp \\ Eggplant\end{tabular} &
 \begin{tabular}[c]{@{}c@{}} Success\end{tabular} \\ \midrule
 RT-1-X~\cite{RT123} & OXE & 0.167 & 0.000 & 0.208 & 0.042 & 0.083 & 0.000 & 0.000 & 0.000 & 0.011 \\
 Octo-Base~\cite{Octo24} & OXE & 0.347 & 0.125 & \cellcolor{linecolor1}{\underline{0.528}} & 0.083 & 0.319 & 0.000 & 0.667 & 0.431 & 0.160 \\
 Octo-Small~\cite{Octo24} & OXE & \cellcolor{linecolor2}{\textbf{0.778}} & \cellcolor{linecolor1}{\underline{0.472}} & 0.278 & 0.097 & 0.403 & 0.042 & \cellcolor{linecolor1}{\underline{0.875}} & 0.569 & 0.300 \\
 OpenVLA~\cite{OpenVLA24} & OXE & 0.041 & 0.000 & 0.333 & 0.000 & 0.125 & 0.000 & 0.083 & 0.041 & 0.010 \\
 RoboVLM~\cite{robovlm25} & OXE & 0.375 & 0.208 & 0.333 & \cellcolor{linecolor1}{\underline{0.250}} & 0.083 & 0.083 & 0.000 & 0.000 & 0.135 \\
 RoboVLM~\cite{robovlm25} & Bridge & 0.542 & 0.292 & 0.250 & 0.250 & 0.458 & 0.125 & 0.583 & \cellcolor{linecolor1}{\underline{0.583}} & 0.313 \\
 SpatialVLA~\cite{spatialvla25} & OXE & 0.250 & 0.208 & 0.417 & 0.208 & 0.583 & 0.250 & 0.792 & 0.708 & 0.344 \\
 SpatialVLA~\cite{spatialvla25} & Bridge & 0.208 & 0.167 & 0.292 & 0.250 & \cellcolor{linecolor1}{\underline{0.625}} & \cellcolor{linecolor1}{\underline{0.292}} & \cellcolor{linecolor2}{\textbf{1.000}} & \cellcolor{linecolor2}{\textbf{1.000}} & \cellcolor{linecolor1}{\underline{0.427}} \\
 \midrule
 \textbf{\ours} & \textbf{Zero-Shot} & \cellcolor{linecolor1}{\underline{0.625}} & \cellcolor{linecolor2}{\textbf{0.583}} & \cellcolor{linecolor2}{\textbf{0.750}} & \cellcolor{linecolor2}{\textbf{0.667}} & \cellcolor{linecolor2}{\textbf{0.917}} & \cellcolor{linecolor2}{\textbf{0.708}} & 0.667 & 0.375 & \cellcolor{linecolor2}{\textbf{0.583}} \\
\bottomrule
\end{tabular}
\vspace{-5pt}
}
\label{tab:widowx}
\end{table*}

\subsection{Simulation Object Manipulation Evaluation on SIMPLER~\cite{simplerenv24}}
We conduct quantitative evaluations of \ours's zero-shot execution performance on Google Robot tasks \& Widow-X tasks and compare it to baselines including Octo~\cite{Octo24}, OpenVLA~\cite{OpenVLA24} and more concurrent works~\cite{robovlm25,spatialvla25}.
The robot follows the planned trajectory generated by the planning module, as described in Sec. \ref{execution}, to execute the task.
Furthermore, leveraging the error detection and re-planning capabilities of VLMs~\cite{GPT4o24,gemini23}, we can make multiple attempts following a single-step execution failure to approximately achieve a closed-loop effect. For fairness, we limit the maximum number of attempts to three. Detailed visualizations and analyses are provided in the~\cref{app:close_loop}.
As shown in \cref{tab:simpler_env,tab:widowx}, despite the training data for Octo and OpenVLA including Google Robot tasks, \ours~demonstrates superior zero-shot performance compared to most baselines.

\subsection{Orientation-Aware Robotic Navigation}
\begin{figure*}[t!]
\begin{center}
\vspace{-5pt}
\includegraphics[width=\linewidth]{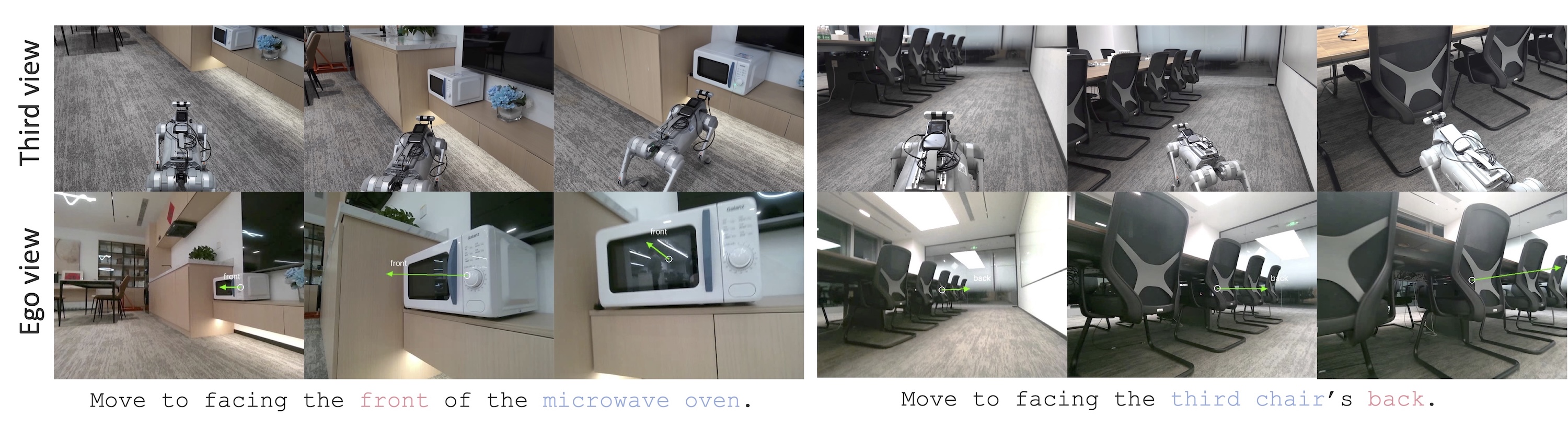}
\vspace{-15pt}
\caption{\textbf{Real-world orientation-aware navigation.} We present both the third-person view and the egocentric view, annotating the predicted orientation of the interacted objects.}
\vspace{-10pt}
\label{fig:navigation}
\end{center}
\end{figure*}

In navigation tasks, reaching an object from its functional side is crucial for subsequent manipulation—for example, approaching a microwave from the front to open its door. To support such scenarios, we extend \textit{semantic orientation} to the navigation domain. As shown in \cref{fig:navigation}, a quadruped robot is tasked with reaching both the correct position and the appropriate facing direction. This orientation-aware constraint enhances the navigation process by ensuring precise alignment with the desired orientation, thereby improving task performance in scenarios where directionality is critical.

\begin{wraptable}{r}{0.505\textwidth}
\vspace{-14pt}
\centering
\setlength{\tabcolsep}{1.5pt}
\caption{\textbf{Spatial reasoning evaluation} on 6-DoF SpatialBench. \textit{Depth-Esti}: Use depth estimation methods such as Metric3D~\cite{Metric3D23} or Moge~\cite{moge24}.}
\vspace{-3pt}
\label{tab:spatial_vqa}
\resizebox{0.485\textwidth}{!}{
\begin{tabular}{lcccccc}
\toprule[0.95pt]
\multirow{2}{*}[-0.5ex]{Method} & 
\multirow{2}{*}[-0.5ex]{Depth-Esti} & \multicolumn{2}{c}{\textbf{Position}} & \multicolumn{2}{c}{\textbf{Orientation}} & \multirow{2}{*}[-0.5ex]{Total} \\
\cmidrule(lr){3-6}
& & \texttt{rel.} & \texttt{abs.} & \texttt{rel.} & \texttt{abs.} &\\ 
\midrule[0.6pt]
\multicolumn{7}{c}{\textit{Blind Evaluation with LLMs}}\\
\midrule[0.6pt]
GPT-3.5-Turbo~\cite{GPT3_20} & \ding{\numexpr55\relax} & 24.5 & 24.9 & 26.7 & 27.5 & 25.7\\
GPT-4-Turbo~\cite{GPT4_23} & \ding{\numexpr55\relax} & 27.2 & 27.3 & 29.2 & 27.9 & 27.8\\
\midrule[0.6pt]
\multicolumn{7}{c}{\textit{General VLMs}}\\
\midrule[0.6pt]
LLaVA-1.5~\cite{LLaVA1.523} & \ding{\numexpr55\relax} & 30.9 & 24.5 & 28.3 & 25.8 & 27.2\\
GPT-4o-mini~\cite{GPT4o24} & \ding{\numexpr55\relax} & 33.3 & 26.9 & 32.5 & 23.8 & 31.0\\
% GPT-4V~\cite{GPT4Vision23} & \ding{\numexpr55\relax} & 37.7 & 32.7 & 36.7 & 27.5 & 33.9\\
GPT-4o~\cite{GPT4o24} & \ding{\numexpr55\relax} & 49.4 & 28.4 & 44.2 & 25.8 & 36.2\\
\midrule[0.6pt]
\multicolumn{7}{c}{\textit{VLMs with Spatial Awareness}}\\
\midrule[0.6pt]
SpaceLLaVA~\cite{SpatialVLM24} & \ding{\numexpr55\relax} & 32.4 & 30.5 & 30.9 & 24.9 & 28.2\\
SpaceMantis~\cite{SpatialVLM24} & \ding{\numexpr55\relax} & 33.6 & 29.2 & 27.2 & 25.0 & 28.9\\
SpatialBot~\cite{SpatialBot24} & \ding{\numexpr51\relax} & 50.9 & 21.6 & 39.6 & 22.9 & 32.7\\
\rowcolor{linecolor1}RoboPoint~\cite{RoboPoint24} & \ding{\numexpr55\relax} & 43.8 & 30.8 & 33.8 & 25.8 & 33.5\\
\rowcolor{linecolor2}\textbf{\ours} & \ding{\numexpr51\relax} & \textbf{59.6} & \textbf{33.8} & \textbf{54.6} & \textbf{31.3} & \textbf{43.9}\\
\bottomrule[0.95pt]
\end{tabular}
}
\vspace{-10pt}
\end{wraptable}

\subsection{Spatial Reasoning Evaluation on 6-DoF SpatialBench}
To assess spatial understanding with full 6-DoF awareness, we introduce \textbf{6-DoF SpatialBench}, a VQA benchmark designed to evaluate both positional and orientational comprehension. Unlike prior benchmarks~\cite{SpatialRGPT24,SpatialBot24,embspatial24,space3d24} that primarily emphasize coarse positional reasoning (\eg, ``to the left,'' ``nearest'') and often overlook orientation or rely on relative metrics, we provide a more fine-grained evaluation with quantitative annotations. It consists of 223 human-annotated samples, each containing an RGB image and a multiple-choice question with 4 options. The benchmark includes two tracks: position and orientation, covering tasks such as object counting, spatial relations, and object-facing direction. All questions and ground-truth answers are curated through \textbf{human annotation}.
We evaluate \ours~on 6-DoF SpatialBench against several VLMs and comparable methods as baselines, as presented in \cref{tab:spatial_vqa}. \ours~consistently outperforms other methods across both tracks, achieving over 18\% improvement.

\section{Limitations \& Conclusions}
\label{sec:limitation}
One notable limitation for decoupled systems like \sofar is that the execution may fail due to a sub-module error, as shown in \cref{app:error},
\ie, robots may place target objects with an error transformation because of unstable grasping or inaccurate visual perception.
For example, the pen will be placed in an unexpected pose due to the rotation during execution.
Future works include integrating scalable data and more advanced models and exploring the potential of combining end-to-end and such decoupled methods, and expanding \sofar to more applications.

We propose \textit{semantic orientation}, a language-grounded representation that links object orientations with intuitive descriptors (\eg, ``plug-in direction''), bridging geometric reasoning and functional semantics. To support this, we construct OrienText300K, a large-scale dataset of 3D models with semantic orientation annotations. Our PointSO model, integrated within the \sofar~system, demonstrates strong performance in both simulated and real-world robotic manipulation tasks.

{
\small\bibliographystyle{iclr2024}
\bibliography{main}
}

%%%%%%%%%%%%%%%%%%%%%%%%%%%%%%%%%%%%%%%%%%%%%%%%%%%%%%%%%%%%

\appendix

\clearpage
\appendix
\section{Robot Setups}\label{app:robot_setup}

\subsection{Simulation Robot Setups}
To ensure fairness, we utilize the same Franka Panda arm for evaluations in both the LIBERO~\cite{LIBERO23} and our Open6DOR V2 benchmarks. For SIMPLER~\cite{simplerenv24}, we use the Google Robot and Widow-X exclusively to conduct the baseline experiments, adhering to all configurations outlined in SIMPLER, as presented in \cref{tab:widowx,tab:simpler_env}. 

\subsection{Real World Robot Setups}
\begin{figure}[h!]
\centering
\includegraphics[width=0.6\linewidth]{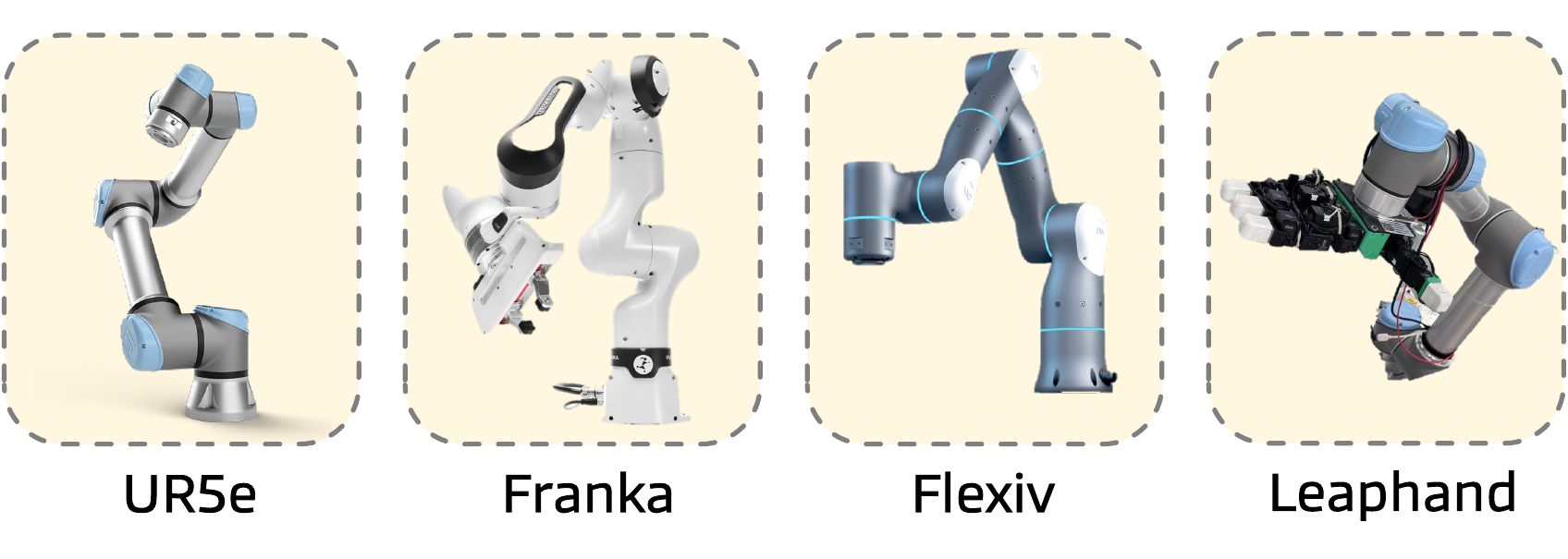}
\vspace{-5pt}
\captionof{figure}{\textbf{The robots used in our real-world experiments.}}
\vspace{-5pt}
\label{fig:robots}
\end{figure}

As for manipulation tasks, in \cref{fig:robots}, we perform 6-DoF rearrangement tasks using the Franka Panda equipped with a gripper and the UR robot arm with a LeapHand, while articulated object manipulation is conducted using the Flexiv arm equipped with a suction tool. All the robot arms mount a RealSense D415 camera at their end for image capturing.

% \begin{wrapfigure}{r}{0.5\textwidth}
%   \centering
%   \vspace{-15pt}
%   \includegraphics[width=0.48\textwidth]{figs/src/robots.pdf}
%   \vspace{-8pt}
%   \caption{\textbf{The robots used in our real-world experiments.}}
%   \label{fig:robots}
%   \vspace{-10pt}
% \end{wrapfigure}

\begin{figure}[h]
\centering
\begin{minipage}[t]{0.48\linewidth}
    \centering
    \includegraphics[width=\linewidth]{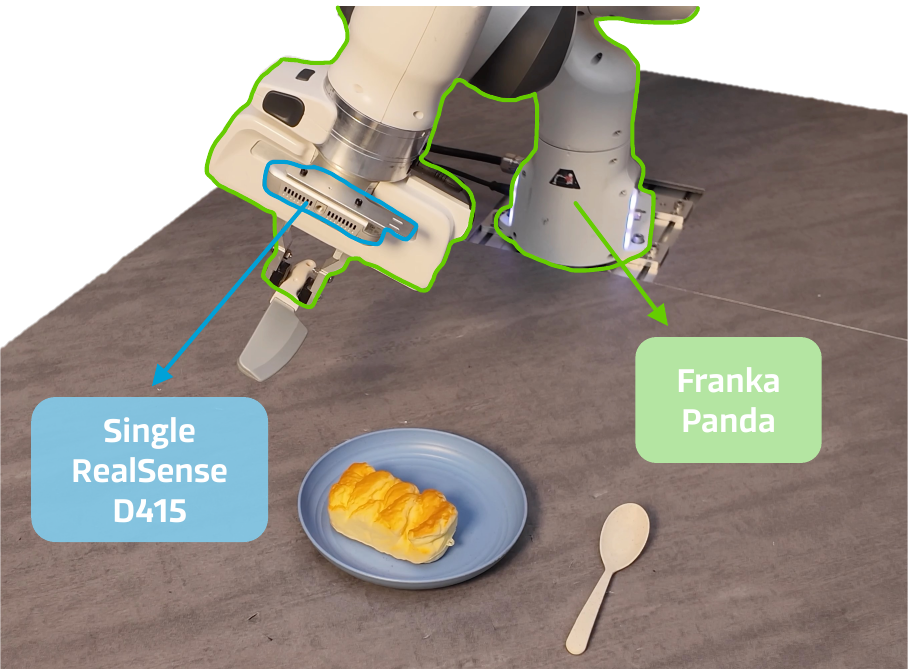}
    \captionof{figure}{\textbf{6-DoF rearrangement robot setup.}}
    \label{fig:franka_setup}
\end{minipage}
\hfill
\begin{minipage}[t]{0.48\linewidth}
    \centering
    \includegraphics[width=\linewidth]{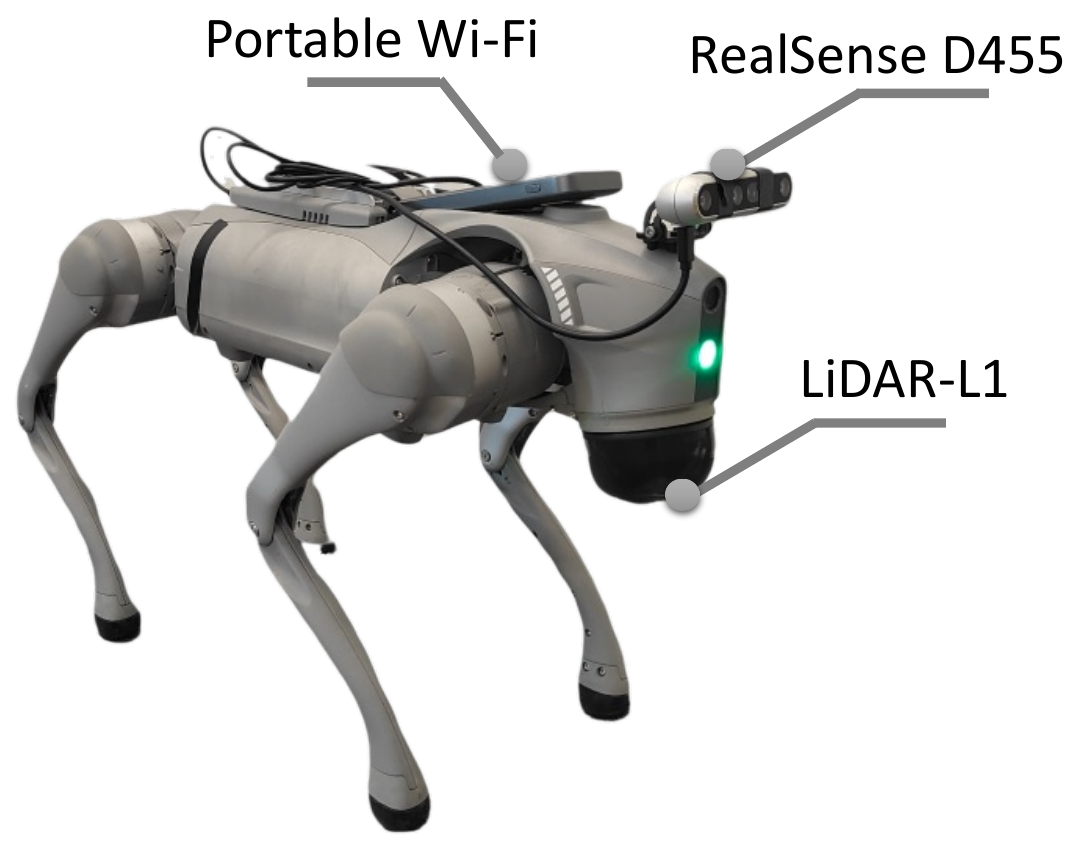}
    \captionof{figure}{\textbf{Navigation robot setup.}}
    \label{fig:dog_setup}
\end{minipage}
% \vspace{-10pt}
\end{figure}

In \cref{fig:franka_setup}, we present the workspace and robotic arm for real-world 6-DoF rearrangement. Unlike Rekep~\cite{ReKep24}, CoPa~\cite{CoPa24} et al., we utilize only a single RealSense D415 camera. This setup significantly reduces the additional overhead associated with environmental setup and multi-camera calibration, and it is more readily reproducible.

As for navigation tasks, we provide a visualization of our robotic dog in~\cref{fig:dog_setup}. Following Uni-Navid~\cite{uninavid24}, our robotic dog is Unitree GO2 and we mount a RealSense D455 camera on the head of the robotic dog. Here, we only use the RGB frames with a resolution of $640\times480$ in the setting of  $90^\circ$ HFOV. We also mount a portable Wi-Fi at the back of the robot dog, which is used to communicate with the remote server (send captured images and receive commands). Unitree GO2 is integrated with a LiDAR-L1, which is only used for local motion planning. 
% \begin{figure}[h!]
% \centering
% \includegraphics[width=0.6\linewidth]{figs/src/franka_setup.pdf}
% \captionof{figure}{\textbf{6-DoF rearrangement robot setup.}}
% \vspace{-10pt}
% \label{fig:franka_setup}
% \end{figure}

\section{Additional Experiments}\label{app:add_exp}
\subsection{Articulated Objects Manipulation Evaluation}
We further integrate \ours~with articulated object manipulation, as illustrated in \cref{tab:manip}, and evaluate its practicality in robotic manipulation tasks using the PartNet-Mobility Dataset within the SAPIEN~\cite{SAPIEN20} simulator. Our experimental setup follows ManipLLM~\cite{ManipLLM24}, employing the same evaluation metrics. Specifically, we directly utilize the segmentation centers provided by SAM as contact points, leverage PointSO to generate contact directions, and use VLM to determine subsequent motion directions. The results demonstrate significant improvements over the baseline. Notably, our model achieves this performance without dividing the data into training and testing sets, operating instead in a fully zero-shot across most tasks. This underscores the robustness and generalization of our approach.
\begin{table*}[t!]
\begin{center}
\small
\caption{\textbf{Zeroshot articulate object manipulation evaluation} within the SAPIEN~\cite{SAPIEN20} simulator using PartNet-Mobility Dataset. Notably, while the baseline methods use distinct training and testing splits, our model achieves robust performance without fine-tuning on the SAPIEN samples.} 
\resizebox{1.0\linewidth}{!}{
\setlength{\tabcolsep}{1.5mm}{
\begin{tabular}{c cc c c c c c c c c c c c c c c}
\toprule[0.95pt]
% \multirow{2}{*}{\textbf{}} &\multirow{2}{*}{\textbf{}} &\multicolumn{15}{c}{\textbf {Train Categories}}\\
\multirow{1}{*}[1.2ex]{Method}
 & \includegraphics[width=0.035\linewidth]{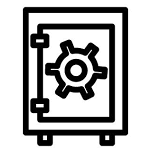}
 &\includegraphics[width=0.035\linewidth]{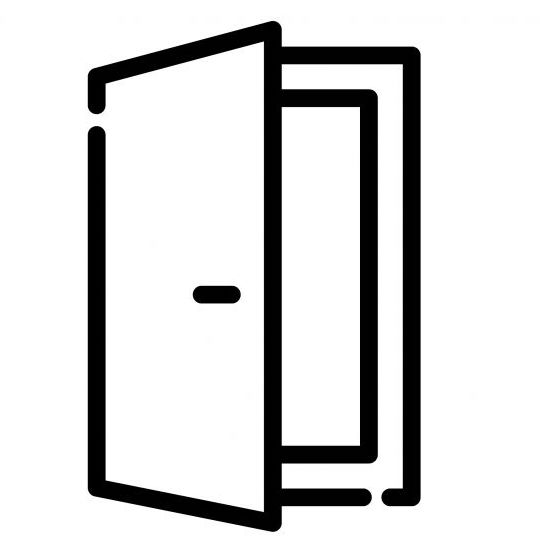}
 &\includegraphics[width=0.035\linewidth]{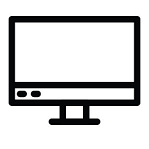}
 &\includegraphics[width=0.035\linewidth]{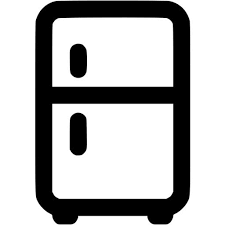}
 &\includegraphics[width=0.035\linewidth]
{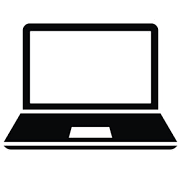}
 &\includegraphics[width=0.035\linewidth]
{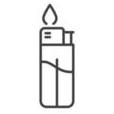}
 &\includegraphics[width=0.035\linewidth]{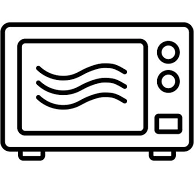}
 &\includegraphics[width=0.035\linewidth]{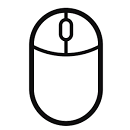}
 &\includegraphics[width=0.035\linewidth]{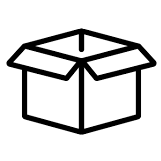}
 &\includegraphics[width=0.035\linewidth]{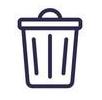}
 &\includegraphics[width=0.035\linewidth]{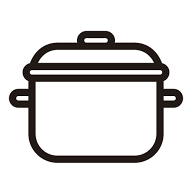}
 &\includegraphics[width=0.035\linewidth]{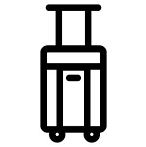}
 &\includegraphics[width=0.035\linewidth]{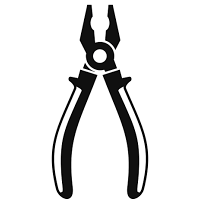}
 &\includegraphics[width=0.035\linewidth]{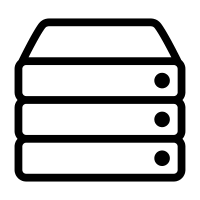}
 &\includegraphics[width=0.035\linewidth]{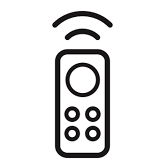}
 &\includegraphics[width=0.035\linewidth]{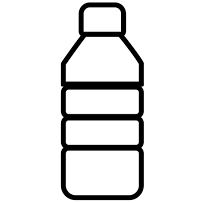}\\
\midrule[0.6pt]
Where2Act~\cite{where2act21} & 0.26 & 0.36 & 0.19 & 0.27 & 0.23 & 0.11 & 0.15 & 0.47 & 0.14 & 0.24 & 0.13 & 0.12 & 0.56 & 0.68 & 0.07 & 0.40\\ %\midrule[0.6pt]
UMPNet~\cite{UMPNet22} & 0.46 & 0.43 & 0.15 & 0.28 & 0.54 & 0.32 & 0.28 & 0.56 & 0.44 & 0.40 &0.10 & 0.23 & 0.18 & 0.54 & 0.20 & 0.42 \\
FlowBot3D~\cite{flowbot3d22} & 0.67 & 0.55 & 0.20 & 0.32 & 0.27 &0.31 & 0.61 & \textbf{0.68} & 0.15 & 0.28 & 0.36 & 0.18 & 0.21 & 0.70 & 0.18 & 0.26\\
Implicit3D~\cite{Implicit3D23} & 0.53 & 0.58 & 0.35 & 0.55 & 0.28 & \textbf{0.66} & 0.58 & 0.51 & 0.52 & 0.57 & 0.45 & 0.34 &0.41 & 0.54 & 0.39 & 0.43\\
ManipLLM ~\cite{ManipLLM24} & 0.68 & 0.64 & 0.36 & 0.77 & 0.43 & 0.62 & 0.65 & 0.61 & 0.65 & 0.52 & 0.53 & \textbf{0.40} & 0.64 & 0.71 & \textbf{0.60} & \textbf{0.64} \\
\midrule[0.6pt] 
\rowcolor{linecolor2}\textbf{\ours} &\textbf{0.75} &\textbf{0.88} & \textbf{0.43} & \textbf{0.85} &\textbf{0.60} & 0.54 & \textbf{0.75} & 0.49 &\textbf{0.58} &\textbf{0.72} & \textbf{0.69} & 0.42 & \textbf{0.70} & \textbf{0.81} & 0.58 & 0.63 \\
\midrule[0.95pt]
% \multirow{2}{*}{\textbf{}} &\multirow{2}{*}{\textbf{}} &\multicolumn{4}{c|}{\textbf {Train Categories} } &
% \multicolumn{10}{c}{\textbf {Test Categories}}\\
\multirow{1}{*}[1.2ex]{Method}
 & \includegraphics[width=0.035\linewidth]{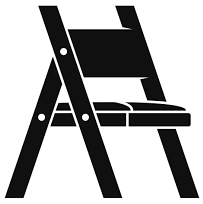}
 &\includegraphics[width=0.035\linewidth]{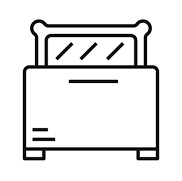}
 &\includegraphics[width=0.035\linewidth]{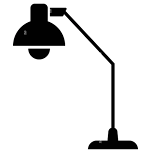}
 &\includegraphics[width=0.035\linewidth]{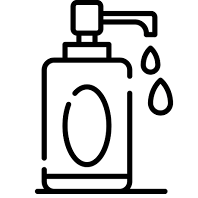}
 &\multicolumn{1}{c|}{\multirow{1}{*}[1.2ex]{{\textbf{AVG}}}}
 &\includegraphics[width=0.035\linewidth]{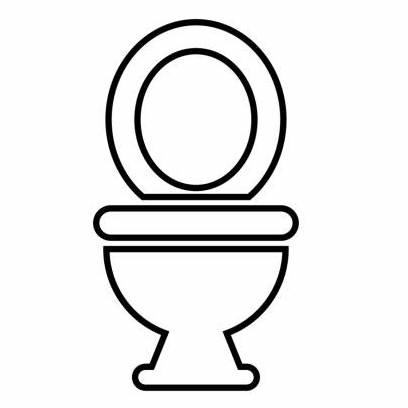}
 &\includegraphics[width=0.035\linewidth]{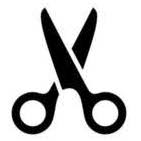}
 &\includegraphics[width=0.035\linewidth]{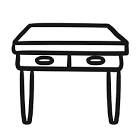}
 &\includegraphics[width=0.035\linewidth]{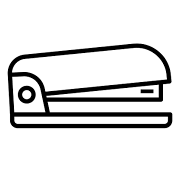}
 &\includegraphics[width=0.035\linewidth]{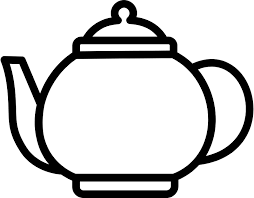}
 &\includegraphics[width=0.035\linewidth]{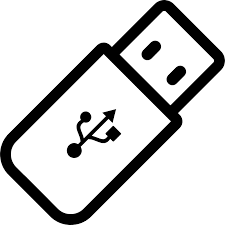}
 &\includegraphics[width=0.035\linewidth]{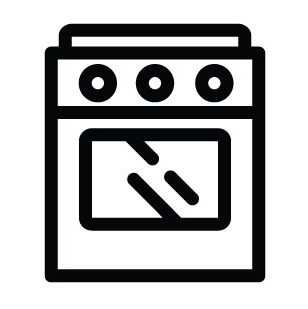}
 &\includegraphics[width=0.035\linewidth]{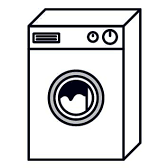}
 &\includegraphics[width=0.035\linewidth]{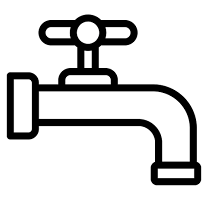}
 &\includegraphics[width=0.035\linewidth]{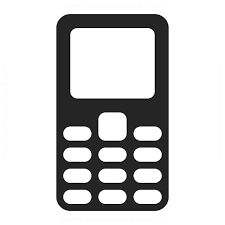}
 &\multirow{1}{*}[1.2ex]{{\textbf{AVG}}}\\
\midrule[0.6pt]
Where2Act~\cite{where2act21} & 0.13 & 0.18 &0.13 & 0.40 & \multicolumn{1}{c|}{0.26} & 0.18 & 0.35 &0.38 & 0.28 & 0.05 & 0.21 & 0.17 & 0.20 & 0.15 &0.15 &0.21 \\
UMPNet~\cite{UMPNet22} & 0.22 & 0.33 & 0.26 & 0.64 & \multicolumn{1}{c|}{0.35} & 0.42 & 0.20 & 0.35 &0.42 & 0.29 & 0.20 & 0.26 & 0.28 & 0.25 & 0.15 & 0.28\\
FlowBot3D~\cite{flowbot3d22} & 0.17 & 0.53 & 0.29 &0.42 & \multicolumn{1}{c|}{0.37} & 0.23 & 0.10 & 0.60 & 0.39 & 0.27 & 0.42 & 0.28 & 0.51 & 0.13 & 0.23 & 0.32
\\
Implicit3D~\cite{Implicit3D23} & 0.27 & 0.65 & 0.20 & 0.33 & \multicolumn{1}{c|}{0.46} & 0.45 & 0.17 & 0.80 & 0.53 &0.15 & 0.69 & 0.41 & 0.31 & 0.30 & 0.31 & 0.41\\
ManipLLM ~\cite{ManipLLM24} & \textbf{0.41} & \textbf{0.75} & 0.44 & 0.67 & \multicolumn{1}{c|}{0.59} & 0.38 &0.22 & 0.81 &\textbf{0.86} & 0.38 & \textbf{0.85} &0.42 &\textbf{0.83} & 0.26 & 0.38 & 0.54 \\ \midrule[0.6pt]
\rowcolor{linecolor2}\textbf{\ours} & 0.35 & 0.68 & \textbf{0.62} & \textbf{0.73} & \multicolumn{1}{c|}{\textbf{0.64}} & \textbf{0.68} & \textbf{0.45} & \textbf{0.90} & 0.77 & \textbf{0.55} & 0.79 & \textbf{0.48} & 0.80 & \textbf{0.56} & \textbf{0.44} & \textbf{0.64}\\
\bottomrule[0.95pt]
\end{tabular}}
\vspace{-10pt}
}
\label{tab:manip} 
\end{center}
\end{table*}

% \begin{table}[h!]
% \centering
% \setlength{\tabcolsep}{9.5pt}
% \caption{Zero-shot performance of LVLMs in EmbSpatial-Bench~\cite{embspatial24}. \textbf{Bold} indicates the best results.}
% \resizebox{0.5\linewidth}{!}{
% \begin{tabular}{lcc}
% \toprule
% Model & Generation & Likelihood \\
% \midrule
% BLIP-2~\cite{BLIP223} & 37.99 & 35.71 \\
% InstructBLIP~\cite{InstructBLIP23} & 38.85 & 33.41 \\
% MiniGPT4~\cite{MiniGPT4_23} & 23.54 & 31.70 \\
% LLaVA-1.6~\cite{LLaVA23} & 35.19 & 38.84 \\
% \midrule
% GPT-4V~\cite{GPT4Vision23} & 36.07 & - \\
% \rowcolor{linecolor1}Qwen-VL-Max~\cite{qwenvl23} & 49.11 & - \\
% \rowcolor{linecolor2}\textbf{\ours} & \textbf{70.88} & - \\
% \bottomrule
% \end{tabular}
% }
% \label{tab:embspatial}
% \end{table}

\begin{wraptable}{r}{0.5\textwidth}
\vspace{-14pt}
\centering
\setlength{\tabcolsep}{10pt}
\caption{Evaluation of EmbSpatial-Bench~\cite{embspatial24}.}
\label{tab:embspatial}
\vspace{-4pt}
\resizebox{0.48\textwidth}{!}{
\begin{tabular}{lcc}
\toprule
Model & Generation & Likelihood \\
\midrule
BLIP-2~\cite{BLIP223} & 37.99 & 35.71 \\
InstructBLIP~\cite{InstructBLIP23} & 38.85 & 33.41 \\
MiniGPT4~\cite{MiniGPT4_23} & 23.54 & 31.70 \\
LLaVA-1.6~\cite{LLaVA23} & 35.19 & 38.84 \\
\midrule
GPT-4V~\cite{GPT4Vision23} & 36.07 & - \\
\rowcolor{linecolor1}Qwen-VL-Max~\cite{qwenvl23} & 49.11 & - \\
\rowcolor{linecolor2}\textbf{\ours} & \textbf{70.88} & - \\
\bottomrule
\end{tabular}
}
\vspace{-10pt}
\end{wraptable}

\subsection{Spatial Reasoning on EmbSpatial-Bench~\cite{embspatial24}}
To further validate the spatial reasoning capabilities of \sofar, we evaluated its performance on the spatial visual-question-answering tasks in EmbSpatial-Bench~\cite{embspatial24}. As reported in \cref{tab:embspatial}, our model substantially outperforms all baseline methods, achieving over a 20\% improvement in overall performance. This result highlights \sofar's effectiveness in spatial understanding and reasoning within complex visual scenes.

\subsection{Cross Embodiment Generalization}
Our approach determines grasp poses by generating masks and plans the target pose and transformation using our PointSO and large language model. It does not rely on trajectory data specific to any robotic arm, making \sofar~embodiment-agnostic. \cref{fig:real_demo} illustrates the diverse embodiments employed in our real-world experiments. Leveraging the GSNet~\cite{GSNet21} algorithm based on LeapHand~\cite{leaphand23}, we perform 6-DoF object manipulation experiments on dexterous hands. We conduct three position-related and three rotation-related experiments. Leveraging the PointSO and large language models, \ours~is capable of performing complex 6-DoF manipulation tasks, such as ``\textit{Upright the fallen wine glass and arrange it neatly in a row with the other wine glasses.}''

\begin{figure*}[t!]
\centering
\vspace{-5pt}
\includegraphics[width=1.0\linewidth]{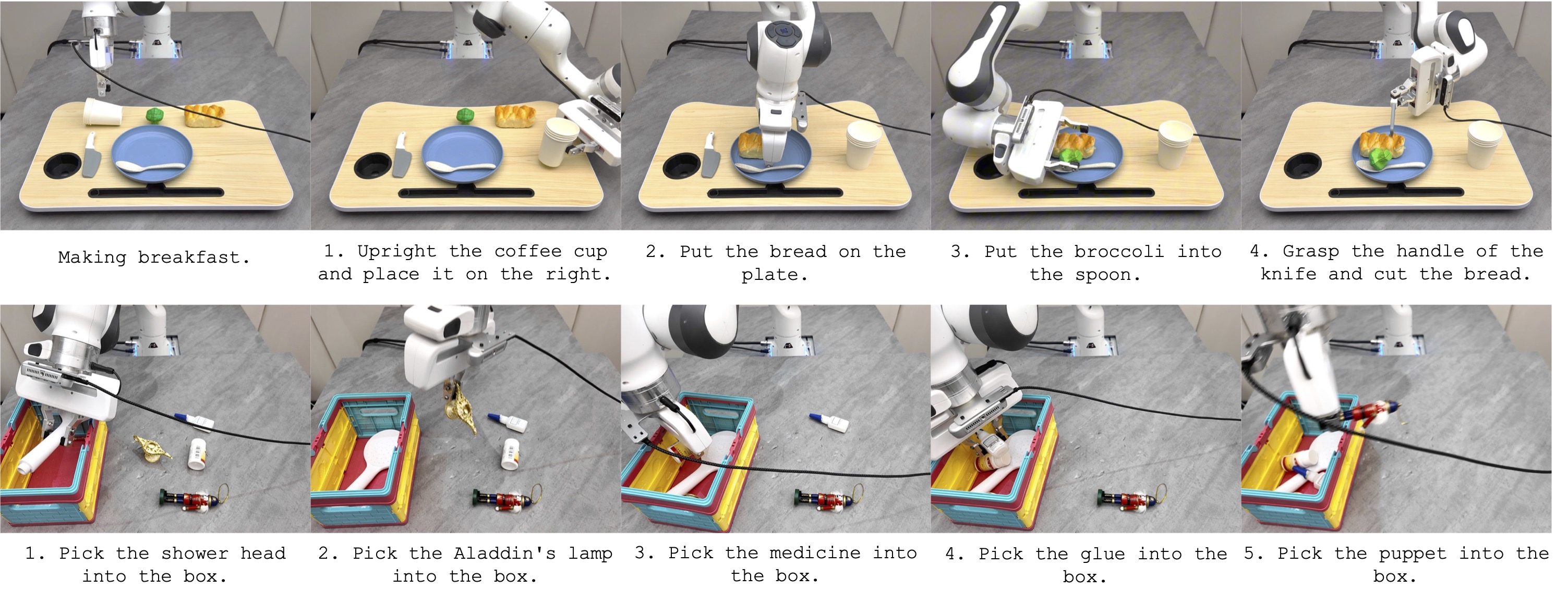}
\vspace{-18pt}
\caption{\textbf{Long-horizon object manipulation} experiment of our \sofar.}
\label{fig:long_horizon}
\vspace{-7pt}
\end{figure*}

\subsection{Long Horizon Object Manipulation Experiment}\label{app:long_horizon}
\cref{fig:long_horizon} illustrates the execution performance of our model on long-horizon tasks. Through the VLM~\cite{GPT4o24,gemini23}, complex instructions such as ``making breakfast'' and ``cleaning up the desktop'' can be decomposed into sub-tasks. In the second example, we deliberately chose uncommon objects as assets, such as ``Aladdin's lamp'' and ``puppets'', but \sofar~is able to successfully complete all tasks.

% \begin{figure}[h!]
% \centering
% \includegraphics[width=0.6\linewidth]{figs/src/close_loop_execution.pdf}
% \vspace{-10pt}
% \captionof{figure}{\textbf{Close-loop execution of our \sofar.}}
% \label{fig:close_loop}
% \end{figure}
\begin{wrapfigure}{r}{0.55\textwidth}
  \centering
  \vspace{-11pt}
  \includegraphics[width=0.53\textwidth]{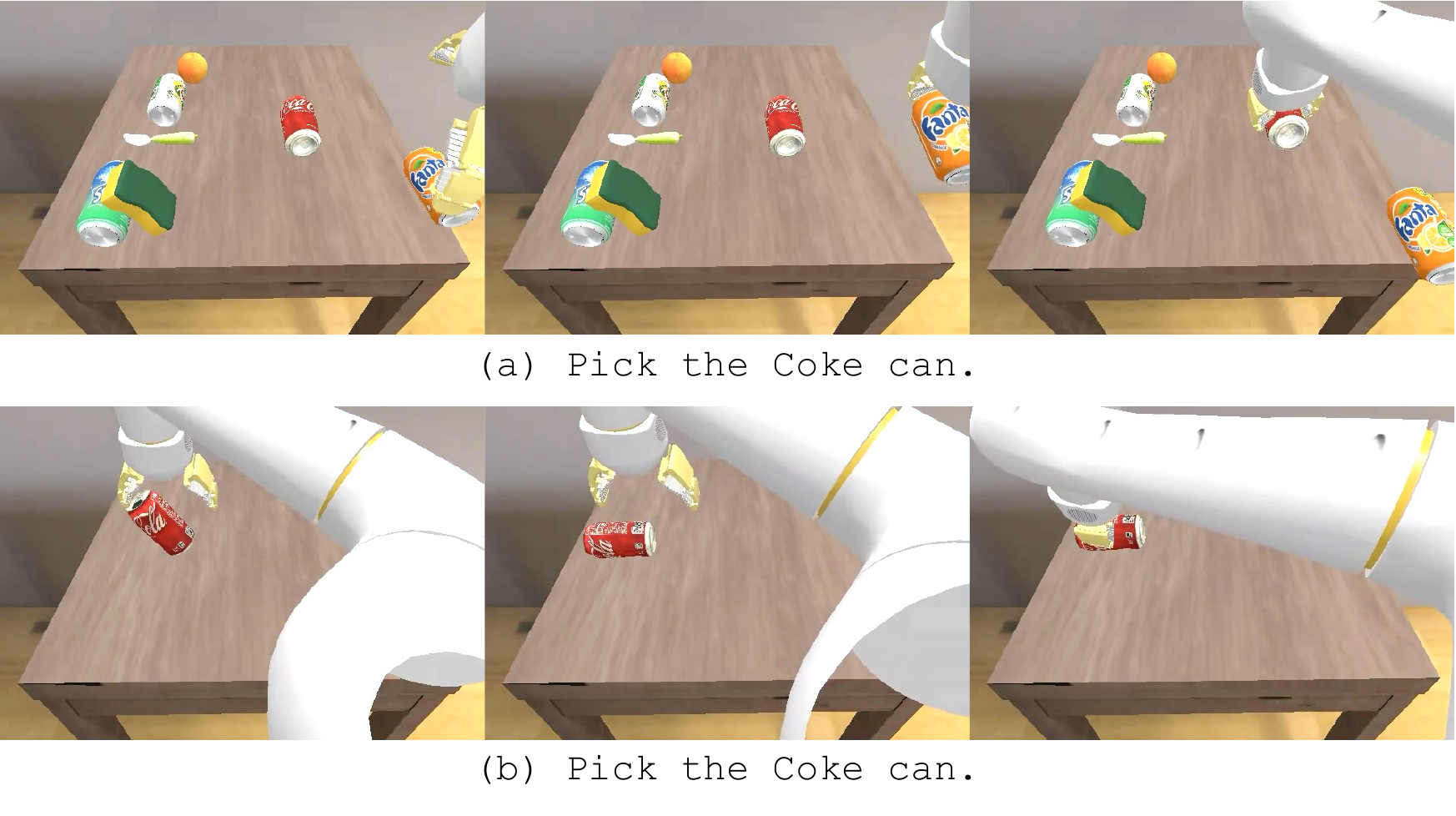}
  \vspace{-5pt}
  \caption{\textbf{Close-loop execution of our \sofar.}}
  \label{fig:close_loop}
  \vspace{-5pt}
\end{wrapfigure}
\subsection{Close-Loop Execution Experiment}\label{app:close_loop}
Similar to ReKep~\cite{ReKep24}, \ours~leverages VLMs~\cite{GPT4o24,gemini23} to perform long-horizon decomposition of complex tasks and employs dual-system VLMs~\cite{GPT4o24,gemini23} to determine the success of execution between tasks and subtasks, enabling closed-loop execution. When a discrepancy between the results and expectations is detected, \ours~re-percepts and re-executes the current subtask.
We demonstrate the closed-loop replan capabilities of \sofar~within Simpler-Env~\cite{simplerenv24} in \cref{fig:close_loop}. The instruction for both tasks is ``pick the coke can'' In \cref{fig:close_loop} (a), the model initially misidentified the coke can as a Fanta can. After correction by the VLM, the model re-identified and located the correct object. In \cref{fig:close_loop} (b), the model accidentally knocks over the Coke can during motion due to erroneous motion planning. Subsequently, the model re-plans and successfully achieves the grasp.

\subsection{In the Wild Evaluation of Semantic Orientation}
We provide a qualitative demonstration of the accuracy of PointSO under in-the-wild conditions, as shown in \cref{fig:in_the_wild}, where the predicted Semantic Orientation is marked in the images. We obtained single-sided point clouds by segmenting objects using Florence-2~\cite{florence2} and SAM~\cite{SAM23} and fed them into PointSO. It can be observed that our model achieves good performance across different views, objects, and instructions, which proves the effectiveness and generalization of PointSO.

\subsection{Cross-View Generalization}
\begin{figure}[t!]
\centering
% \vspace{-5pt}
\begin{minipage}[b]{0.46\linewidth}
    \centering
    \includegraphics[width=\linewidth]{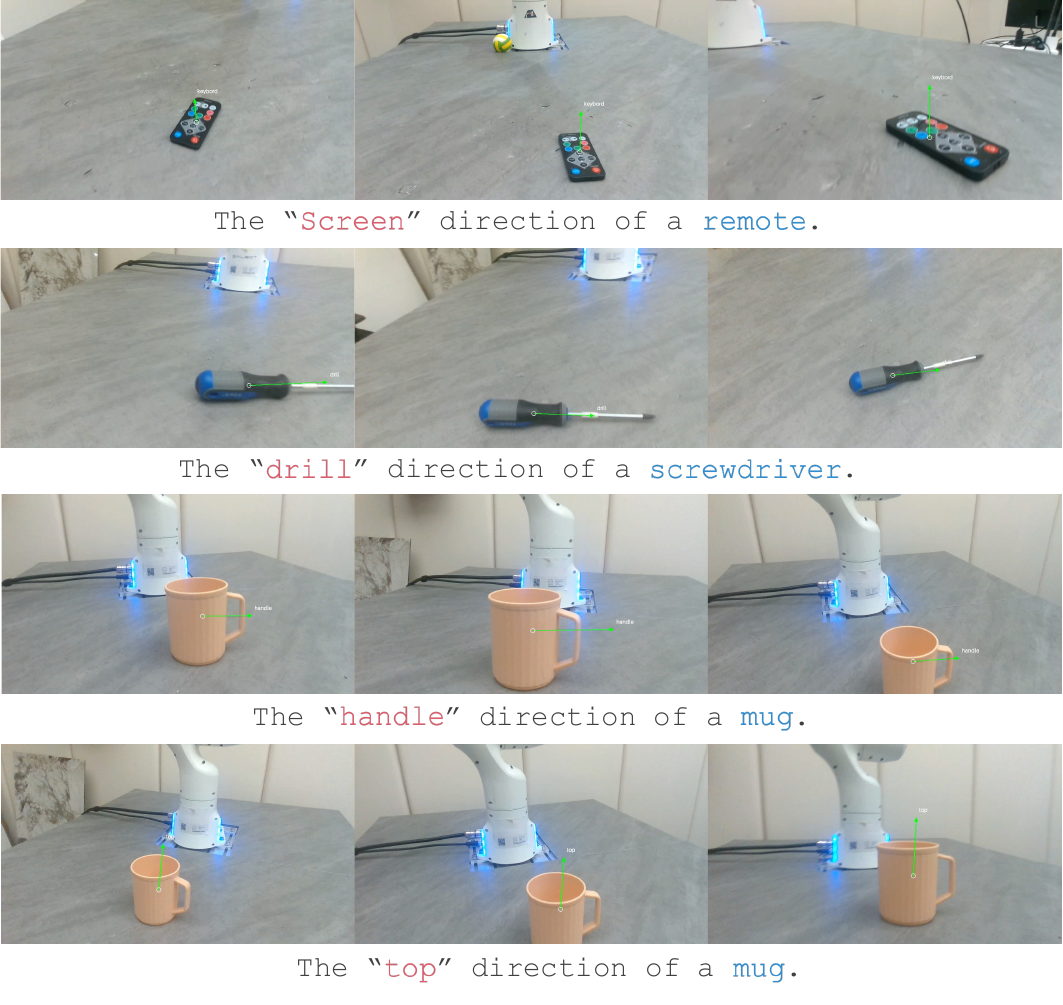}
    \vspace{-15pt}
    \captionof{figure}{\textbf{In-the-wild evaluation} of PointSO.}
    \label{fig:in_the_wild}
\end{minipage}
\hfill
\begin{minipage}[b]{0.52\linewidth}
    \centering
    % \vspace{-1pt}
    \includegraphics[width=\linewidth]{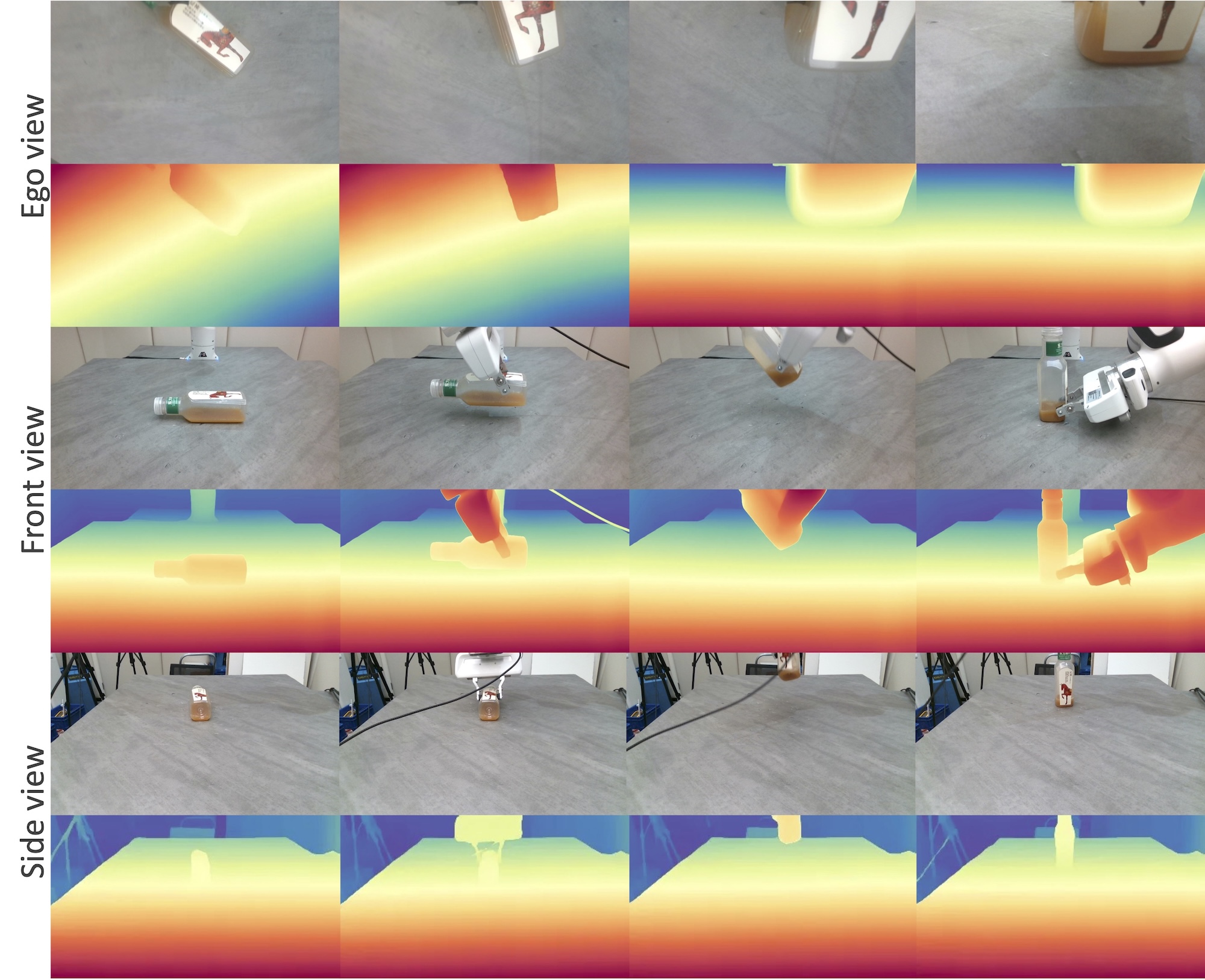}
    % \vspace{-10pt}
    \captionof{figure}{\textbf{Cross view generalization} of \sofar.}
    \label{fig:cross_view}
\end{minipage}
\end{figure}

\sofar~gets point clouds in the world coordinate system using an RGB-D camera to obtain grasping poses, and it is not limited to a fixed camera perspective. In addition, PointSO generates partial point clouds from different perspectives through random camera views to serve as data augmentation for training data, which also generalizes to camera perspectives in the real world. \cref{fig:cross_view} illustrates \sofar's generalization capability for 6-DoF object manipulation across different camera poses. It can be observed that whether it's a front view, side view, or ego view, \sofar~can successfully execute the ``upright the bottle'' instruction.

% \begin{figure}[h!]
% \centering
% \includegraphics[width=0.6\linewidth]{figs/src/in_the_wild.pdf}
% \vspace{-10pt}
% \captionof{figure}{\textbf{In the wild evaluation of PointSO.}}
% \label{fig:in_the_wild}
% \end{figure}

\vspace{-16pt}

\subsection{Failure Case Distribution Analysis}
\label{app:error}
\begin{figure}[h!]
\centering
\includegraphics[width=0.68\linewidth]{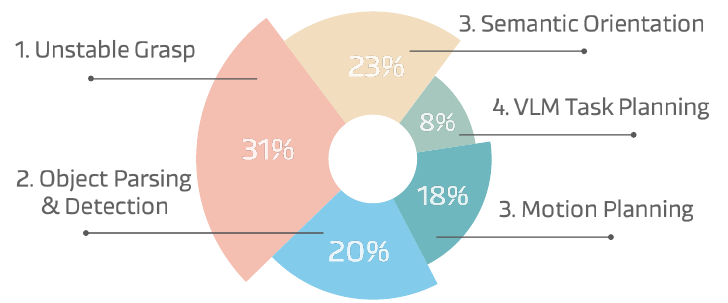}
\captionof{figure}{\textbf{Failure case distribution analysis} of our \sofar.}
\label{fig:failure_case}
\vspace{-5pt}
\end{figure}

Based on the failure cases from real-world experiments, we conducted a quantitative analysis of the failure case distribution for \sofar, with the results shown in \cref{fig:failure_case}. It can be observed that 31\% of the failures originated from grasping issues, including objects being too small, inability to generate reasonable grasping poses, and instability after grasping leading to sliding or dropping. Next, 23\% were due to incorrect or inaccurate Semantic Orientation prediction. For tasks such as upright or upside - down, highly precise angle estimation (<5°) is required for smooth execution. Object analysis and detection accounted for approximately 20\% of the errors. The instability of open-vocabulary detection modules like Florence2~\cite{florence2} and Grounding DINO~\cite{groundingdino23} often led to incorrect detection of out-of-distribution objects or object parts. In addition, since our Motion Planning did not take into account the working space range of the robotic arm and potential collisions of the manipulated object, occasional deadlocks and collisions occurred during motion. Finally, there were issues with the Task Planning of the VLM~\cite{GPT4o24,gemini23}. For some complex Orientations, the VLM occasionally failed to infer the required angles and directions to complete the task. Employing a more powerful, thought-enabled VLM~\cite{gpt_o1} might alleviate such errors.

\section{Ablation Study}\label{app:ablation}

\subsection{Semantic Orientation Ablation}
To demonstrate that the proposed semantic orientation indeed plays a crucial role in robotic tasks—rather than the observed effects being attributable to other factors such as Chain-of-Thought reasoning—we conduct conduct ablation experiments for methodological differences between the baselines, including whether to add semantic orientation in the scene graph and whether to use CoT, as shown in Tab. \ref{tab:without_orientation}.
\begin{table}[t!]
\setlength{\tabcolsep}{4pt}
\caption{\textbf{Ablation study of composition module} of \sofar.
}
\vspace{3pt}
\label{tab:without_orientation}
\centering
\resizebox{1.0\linewidth}{!}{
\begin{tabular}{cccccccccccc}
\toprule
\multirow{2}{*}[-0.5ex]{CoT} & \multirow{2}{*}[-0.5ex]{Orient.} & \multicolumn{3}{c}{\textbf{Position Track}} & \multicolumn{4}{c}{\textbf{Rotation Track}} & \multicolumn{3}{c}{\textbf{6-DoF Track}}\\
\cmidrule(lr){3-5} \cmidrule(lr){6-9} \cmidrule(lr){10-12}
& & Level\,0 & Level\,1 & Overall & Level\,0 & Level\,1 & Level\,2 & Overall & Position & Rotation & Overall \\ 
\midrule
\ding{\numexpr55\relax} & \ding{\numexpr55\relax} & 95.4 & 77.7 & 91.9 & 17.2 & 8.4 & 11.4 & 13.0 & 92.7 & 15.5 & 14.2\\
\ding{\numexpr51\relax} & \ding{\numexpr55\relax} & \textbf{96.3} & \textbf{81.6} & \textbf{93.3} & 16.3 & 8.9 & 11.0 & 12.9 & \textbf{93.0} & 15.1 & 13.7\\
\ding{\numexpr55\relax} & \ding{\numexpr51\relax} & 95.6 & 77.2 & 91.7 & 63.3 & 35.4 & 61.8 & 52.3 & 92.7 & 48.3 & 45.8\\
\rowcolor{linecolor2}\ding{\numexpr51\relax} & \ding{\numexpr51\relax} & 96.0 & 81.5 & 93.0 & \textbf{68.6} & \textbf{42.2} & \textbf{70.1} & \textbf{57.0} & 92.7 & \textbf{52.7} & \textbf{48.7}\\
\bottomrule[0.6pt]
\end{tabular}
}
\vspace{-4pt}
\end{table}

\begin{wraptable}{r}{0.52\textwidth}
\setlength{\tabcolsep}{5.5pt}
\vspace{-13pt}
\caption{\textbf{Data scaling property} of semantic orientation with different training data scales evaluated on OrienText300K validation split. All experiments are conducted with the PointSO-Base variant.}
\label{tab:scaling_law}
\centering
\vspace{-4pt}
\resizebox{0.5\textwidth}{!}{
\begin{tabular}{lccccc}
\toprule[0.95pt]
    Data Scale & \texttt{45°} & \texttt{30°} & \texttt{15°} & \texttt{5°} & Average \\ 
    \midrule[0.6pt]
    15K & 57.03 & 46.09 & 39.84 & 27.34 & 42.58 \\
    35K & 61.72 & 53.13 & 43.75 & 30.47 & 47.27 \\
    150K & 76.56 & 72.66 & 66.41 & 56.25 & 67.97 \\
    \rowcolor{linecolor2}350K & \textbf{79.69} & \textbf{77.34} & \textbf{70.31} & \textbf{62.50} & \textbf{72.46} \\
\bottomrule[0.95pt]
\end{tabular}
}
\vspace{-10pt}
\end{wraptable}

\subsection{Scaling Law}
The scaling capability of models and data is one of the most critical attributes today and a core feature of foundation models~\cite{FoundationModel21}. We investigate the performance of PointSO across different data scales, as illustrated in \cref{tab:scaling_law}. 
We obtain the subset for OrienText300K from Objaverse-LVIS, which consists of approximately 46,000 3D objects with category annotations. The selection was based on the seven criteria mentioned in the main text. Objects meeting all seven criteria formed the strict subset, comprising around 15k objects. When including objects without textures and those of lower quality, the total increases to approximately 26k objects.
It can be seen that the increase in data volume is the most significant factor driving the performance improvement of PointSO. It can be anticipated that with further data expansion, such as Objaverse-XL~\cite{ObjaverseXL23}, PointSO will achieve better performance.

% \begin{table}[t!]
% \setlength{\tabcolsep}{5.8pt}
% \caption{\textbf{Ablation of multi-modal fusion} in PointSO.
%  All the experiments are under the PointSO-Base variant.}
% \label{tab:fusion}
% \centering
% \resizebox{0.6\linewidth}{!}{
% \begin{tabular}{lccccc}
% \toprule[0.95pt]
% Data Scale & \texttt{45°} & \texttt{30°} & \texttt{15°} & \texttt{5°} & Average \\ 
% \midrule[0.6pt]
% Cross-attention & 74.22 & 70.31 & 63.28 & 57.03 & 66.21 \\
% Multiplication & 74.22 & 69.53 & 60.16 & 56.25 & 65.04 \\
% \rowcolor{linecolor2}Addition & \textbf{79.69} & \textbf{77.34} & \textbf{70.31} & \textbf{62.50} & \textbf{72.46} \\
% Concat & 66.41 & 60.94 & 52.34 & 43.75 & 55.86 \\
% \bottomrule[0.95pt]
% \end{tabular}
% }
% \end{table}
\begin{wraptable}{r}{0.52\textwidth}
\vspace{-13pt}
\setlength{\tabcolsep}{5.2pt}
\caption{\textbf{Ablation study of multi-modal fusion} in PointSO. All experiments are conducted with the PointSO-Base variant.}
\label{tab:fusion}
\centering
\vspace{-4pt}
\resizebox{0.5\textwidth}{!}{
\begin{tabular}{lccccc}
\toprule[0.95pt]
Fusion Method & \texttt{45°} & \texttt{30°} & \texttt{15°} & \texttt{5°} & Avg. \\ 
\midrule[0.6pt]
Cross-attn & 74.22 & 70.31 & 63.28 & 57.03 & 66.21 \\
Multiplication & 74.22 & 69.53 & 60.16 & 56.25 & 65.04 \\
\rowcolor{linecolor2}Addition & \textbf{79.69} & \textbf{77.34} & \textbf{70.31} & \textbf{62.50} & \textbf{72.46} \\
Concat & 66.41 & 60.94 & 52.34 & 43.75 & 55.86 \\
\bottomrule[0.95pt]
\end{tabular}
}
\vspace{-10pt}
\end{wraptable}

\subsection{Cross-Modal Fusion Choices}\label{app:fusion}
We further conduct an ablation study on the multi-modal fusion methods in PointSO, testing commonly used feature fusion techniques such as cross-attention, multiplication, addition, and concatenation, as shown in \cref{tab:fusion}. The results indicate that simple addition achieves the best performance. This may be attributed to the fact that instructions in the semantic domain are typically composed of short phrases or sentences, and the text CLS token already encodes sufficiently high-level semantic information.

\begin{table*}[h!]
\setlength{\tabcolsep}{2.1pt}
\caption{\textbf{Ablation study of open vocabulary detection modules} on Open6DOR perception tasks.
}
\vspace{-3pt}
\label{tab:detection_ab}
\centering
\resizebox{\linewidth}{!}{
\begin{tabular}{lccccccccccc}
\toprule[0.95pt]
\multirow{2}{*}[-0.5ex]{Method} & \multicolumn{3}{c}{\textbf{Position Track}} & \multicolumn{4}{c}{\textbf{Rotation Track}} & \multicolumn{3}{c}{\textbf{6-DoF Track}} & \multirow{2}{*}[-0.5ex]{Time Cost (s)}\\
\cmidrule(lr){2-4} \cmidrule(lr){5-8} \cmidrule(lr){9-11}
& Level 0 & Level 1 & Overall & Level 0 & Level 1 & Level 2 & Overall & Position & Rotation & Overall \\ 
\midrule[0.6pt]
YOLO-World~\cite{yoloworld24} & 59.0 & 37.7 & 53.3 & 48.3 & 36.1 & 62.0 & 44.9 & 53.4 & 44.6 & 27.8 & \textbf{7.4s}\\
\rowcolor{linecolor1}Grounding DINO~\cite{groundingdino23} & 92.2 & 71.5 & 86.7 & 64.7 & 41.1 & 69.8 & 55.5 & 87.2 & 51.6 & 44.6 & 9.2s\\
\rowcolor{linecolor2}Florence-2~\cite{florence2} & \textbf{96.0} & \textbf{81.5} & \textbf{93.0} & \textbf{68.6} & \textbf{42.2} & \textbf{70.1} & \textbf{57.0} & \textbf{92.7} & \textbf{52.7} & \textbf{48.7} & 8.5s\\
\bottomrule[0.95pt]
\end{tabular}
\vspace{-5pt}
}
\end{table*}

\subsection{Open Vocabulary Object Detection Module}
\sofar~utilize an open vocabulary detection foundation model to localize the interacted objects or parts, then generate masks with SAM~\cite{SAM23}. Although not the SOTA performance on the COCO benchmark, Florence-2~\cite{florence2} exhibits remarkable generalization in in-the-wild detection tasks, even in simulator scenarios. \cref{tab:detection_ab} illustrates the performance of various detection modules in Open6DOR~\cite{Open6DOR24} Perception, where Florence-2 achieves the best results and outperforms Grounding DINO~\cite{groundingdino23} and YOLO-World~\cite{yoloworld24}.

\section{Additional Implementation Details}\label{app:implementation_details}

\subsection{Detail Real World Experiment Results}\label{app:detail_realworld}
\begin{figure}[t!]
\centering
\includegraphics[width=0.7\linewidth]{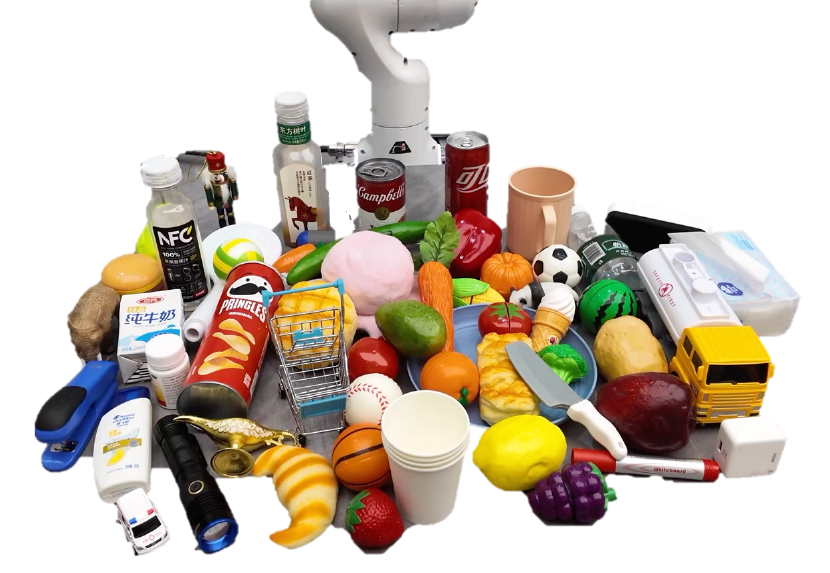}
\vspace{-5pt}
\captionof{figure}{\textbf{The real-world assets used in our real-world experiments}. More than 100 diverse objects are used in our 6-DoF rearrangement experiments.}
\label{fig:real_obj}
\vspace{-5pt}
\end{figure}

To fully demonstrate the generalization of \sofar~rather than cherry-picking, we carefully design 60 different real-world experimental tasks, covering more than 100 different and diverse objects. Similar to the Open6DOR~\cite{Open6DOR24} benchmark in the simulator, we divide these 60 tasks into three parts: position-track, orientation-track, and the most challenging comprehensive \& 6-DoF-track. Each track is further divided into simple and hard levels. The position-simple track includes tasks related to front \& back \& left \& right spatial relationships, while the position-hard track includes tasks related to between, center, and customized. The orientation-simple track includes tasks related to the orientation of object parts, and the orientation-hard track includes tasks related to whether the object is upright or flipped (with very strict requirements for angles in both upright and flipped cases). Comprehensive tasks involve complex instruction understanding and long-horizon tasks; 6-DoF tasks simultaneously include requirements for both object position and orientation instructions. In \cref{tab:detailed_realworld}, we present the complete task instructions, as well as the performance metrics of \sofar~and the baseline. Due to the large number of tasks, we performed each task three times. It can be seen that \sofar~achieves the best performance in all tracks, especially in the orientation-track and comprehensive \& 6-DoF-track. We also show all the objects used in the real-world experiments in \cref{fig:real_obj}, covering a wide range of commonly and uncommonly used objects in daily life.

\begin{table*}[t!]
\setlength{\tabcolsep}{2.8pt}
\caption{\textbf{Detailed zero-shot real-world 6-DoF rearrangement results}.}
\label{tab:detailed_realworld}
\centering
\resizebox{\linewidth}{!}{
\begin{tabular}{lcccc}
\toprule[0.95pt]
    Task & CoPa~\cite{CoPa24} & ReKep-Auto~\cite{ReKep24} & \sofar-LLaVA~(Ours) & \sofar~(Ours) \\ 
    \midrule[0.6pt]
    \multicolumn{5}{c}{\textit{Positional Object Manipulation}}\\
    \midrule
    Move the soccer ball to the right of the bread. & 2/3 & 3/3 & 3/3 & \textbf{3/3} \\
    Place the doll to the right of the lemon. & 3/3 & 3/3 & 3/3 & \textbf{3/3} \\
    Put the pliers on the right side of the soccer ball. & 1/3 & 1/3 & 3/3 & \textbf{2/3} \\
    Move the pen to the right of the doll. & 3/3 & 2/3 & 3/3 & \textbf{3/3} \\
    Place the carrot on the left of the croissant. & 2/3 & 3/3 & 2/3 & \textbf{2/3} \\
    Move the avocado to the left of the baseball. & 3/3 & 2/3 & 2/3 & \textbf{3/3} \\
    Pick the pepper and place it to the left of the charger. & 1/3 & 2/3 & 2/3 & \textbf{2/3} \\
    Place the baseball on the left side of the mug. & 3/3 & 2/3 & 2/3 & \textbf{3/3} \\
    Arrange the flower in front of the potato. & 2/3 & 3/3 & 2/3 & \textbf{3/3} \\
    Put the volleyball in front of the knife. & 3/3 & 3/3 & 3/3 & \textbf{3/3} \\
    Place the ice cream cone in front of the potato. & 2/3 & 3/3 & 2/3 & \textbf{3/3} \\
    Move the bitter melon to the front of the forklift. & 2/3 & 1/3 & 2/3 & \textbf{2/3} \\
    Place the orange at the back of the stapler. & 3/3 & 2/3 & 3/3 & \textbf{3/3} \\
    Move the panda toy to the back of the shampoo bottle. & 2/3 & 3/3 & 3/3 & \textbf{2/3} \\
    pick the pumpkin and place it behind the pomegranate. & \textbf{3/3} & 2/3 & 1/3 & 2/3 \\
    Place the basketball at the back of the board wipe. & 2/3 & 2/3 & 3/3 & \textbf{2/3} \\
    Put the apple inside the box. & 3/3 & 2/3 & 3/3 & \textbf{3/3} \\
    Place the waffles on the center of the plate. & 3/3 & 2/3 & 3/3 & \textbf{3/3} \\
    Move the hamburger into the bowl.& 2/3 & 2/3 & 2/3 & \textbf{3/3} \\
    Pick the puppet and put it into the basket. & 1/3 & 2/3 & 2/3 & \textbf{2/3} \\
    Drop the grape into the box. & 2/3 & 3/3 & 3/3 & \textbf{2/3} \\
    Put the doll between the lemon and the USB. & 2/3 & 2/3 & 2/3 & \textbf{3/3} \\
    Set the duck toy in the center of the cart, bowl, and camera. & 2/3 & 1/3 & 2/3 & \textbf{2/3} \\
    Place the strawberry between the Coke bottle and the glue. & 2/3 & 2/3 & 3/3 & \textbf{3/3} \\
    Put the pen behind the basketball and in front of the vase. & 2/3 & 1/3 & 2/3 & \textbf{2/3} \\
    Total success rate& 74.7\% & 72.0\% & 81.3\% & \textbf{85.3\%} \\
    \midrule
    \multicolumn{5}{c}{\textit{Orientational Object Manipulation}}\\
    \midrule
    Turn the yellow head of the toy car to the right. & 2/3 & 2/3 & 1/3 & \textbf{2/3} \\
    Adjust the knife handle so it points to the right. & 2/3 & 1/3 & 2/3 & \textbf{2/3} \\
    Rotate the cap of the bottle towards the right. & 2/3 & 2/3 & 2/3 & \textbf{2/3}\\
    Rotate the tip of the screwdriver to face the right. & 0/3 & 0/3 & 1/3 & \textbf{1/3}\\
    Rotate the stem of the apple to the right. & 0/3 & 1/3 & 1/3 & \textbf{2/3}\\
    Turn the front of the toy car to the left. & 0/3 & 0/3 & 2/3 & \textbf{2/3} \\
    Rotate the cap of the bottle towards the left. & 2/3 & 1/3 & 1/3 & \textbf{2/3}\\
    Adjust the pear's stem to the right. & 1/3 & 1/3 & 1/3 & \textbf{1/3}\\
    Turn the mug handle to the right. & 1/3 & 1/3 & 2/3 & \textbf{2/3}\\
    Rotate the handle of the mug to towards right.& 2/3 & 1/3 &\textbf{2/3} & 1/3\\
    Rotate the box so the text side faces forward. & 0/3 & 1/3 & 0/3 & \textbf{1/3}\\
    Adjust the USB port to point forward. & 0/3 & 0/3 & 1/3 & \textbf{1/3}\\
    Set the bottle upright. & 0/3 & 1/3 & 0/3 & \textbf{1/3}\\
    Place the coffee cup in an upright position. & 1/3 & 1/3 & 2/3 & \textbf{2/3}\\
    Upright the statue of liberty& 0/3 & 0/3 & \textbf{1/3} & 0/3\\
    Stand the doll upright. & 0/3 & 1/3 & 0/3 & \textbf{1/3}\\
    Right the Coke can. & 0/3 & 0/3 & 1/3 & \textbf{1/3}\\
    Flip the bottle upside down. & 0/3 & 0/3 & 0/3 & \textbf{1/3}\\
    Turn the coffee cup upside down. & 0/3 & 0/3 & 1/3 & \textbf{1/3}\\
    Invert the shampoo bottle upside down. & 0/3 & 0/3 & 0/3 & \textbf{0/3}\\
    Total success rate& 21.7\% & 23.3\% & 35.0\% & \textbf{43.3\%} \\
    \midrule
    \multicolumn{5}{c}{\textit{Comprehensive 6-DoF Object Manipulation}}\\
    \midrule
    Pull out a tissue.& 3/3 & 3/3 & 2/3 & \textbf{3/3}\\
    Place the right bottle into the
    box and arrange it in a 3×3 pattern. & 0/3 & 0/3 & 0/3 & \textbf{1/3}\\
    Take the tallest box and position it on the right side. & 1/3 & 1/3 & 3/3 & \textbf{3/3}\\
    Grasp the error bottle and put it on the right side. & 1/3 & 2/3 & 1/3 & \textbf{2/3} \\
    Take out the green test tube and place it between the two bottles. & 2/3 & 2/3 & 3/3 & \textbf{3/3}\\
    Pack the objects on the table into the box one by one. & 1/3 & 1/3 & 0/3 & \textbf{1/3}\\
    Rotate the loopy doll to face the yellow dragon doll & 0/3 & 1/3 & 1/3 & \textbf{1/3}\\
    Right the fallen wine glass and arrange it neatly in a row. & 0/3 & 0/3 & 0/3 & \textbf{0/3}\\
    Grasp the handle of the knife and cut the bread.& 0/3 & 0/3 & 0/3 & \textbf{1/3}\\
    Pick the baseball into the cart and turn the cart to facing right. & 0/3 & 0/3 & 1/3 & \textbf{2/3}\\
    Place the mug on the left of the ball and the handle turn right. & 0/3 & 0/3 & 1/3 & \textbf{1/3}\\
    Aim the camera at the toy truck. & 1/3 & 0/3 & 1/3 & \textbf{1/3}\\
    Rotate the flashlight to illuminate the loopy. & 0/3 & 0/3 & 1/3 & \textbf{1/3}\\
    Put the pen into the pen container. & 0/3 & 1/3 & 0/3 & \textbf{1/3} \\
    Pour out chips from the chips cylinder to the plate. & 0/3 & 1/3 & 1/3 & \textbf{1/3} \\
    Total success rate& 20.0\% & 26.7\% & 33.3\% & \textbf{48.9\%} \\
    \bottomrule[0.95pt]
\end{tabular}
\vspace{-13pt}
}
\end{table*}
\vspace{-20pt}

\begin{center}
\begin{table*}[t!]
\caption{\textbf{Training recipes for PointSO and \ours-LLaVA}.}
\label{tab:hyper_params}
\vspace{-3pt}
\centering
\resizebox{1.0\linewidth}{!}{
\begin{tabular}{lccccc}
 & \multicolumn{4}{c}{\textbf{PointSO}} & \multicolumn{1}{c}{\textbf{\ours-LLaVA}}\\
 \toprule[0.95pt]
 Config & Small & Base & Large & Finetune & SFT\\
 \midrule[0.6pt]
 optimizer & AdamW & AdamW & AdamW & AdamW & AdamW\\
 learning rate & 5e-5 & 5e-5 & 2e-5 & 5e-5 & 2e-5 \\
 weight decay & 5e-2 & 5e-2 & 5e-2 & 5e-2 & 0 \\
 learning rate scheduler & cosine & cosine & cosine & cosine & cosine \\
 training epochs & 300 & 300 & 300 & 50 & 2 \\
 warmup epochs & 10 & 10 & 10 & 5 & 0.03 \\
 batch size & 256 & 256 & 256 & 256 & 128 \\
 drop path rate & 0.2 & 0.2 & 0.2 & 0.2 & - \\
 \midrule[0.6pt]
 number of points & 10000 & 10000 & 10000 & 10000 & - \\
 number of point patches & 512 & 512 & 512 & 512 & - \\
 point patch size & 32 & 32 & 32 & 32 & - \\
 \midrule[0.6pt]
 augmentation & Rot\&Part\&Noise & Rot\&Part\&Noise & Rot\&Part\&Noise & Rotation & - \\
 \midrule[0.6pt]
 GPU device & 8$\times$H800 & 8$\times$H800 & 8$\times$H800 & 8$\times$H800 & 8$\times$H800 \\
\bottomrule[0.95pt]
\end{tabular}
}
\end{table*}
\end{center}

% \begin{table}[!t]
% \setlength\tabcolsep{2.5pt}
% \caption{\textbf{Details of PointSO model variants}. This table format follows~\citet{ViT}.}
% \vspace{-6pt}
% \label{tab:PointSO_configs}
% \begin{center}
% \resizebox{0.65\linewidth}{!}{
% \begin{tabular}{lcccccc}
% \toprule[0.95pt]
% Model & CLIP&  Layers & Hidden size & MLP size & Heads & \#Params\\
% \midrule[0.6pt]
% Small & ViT-B/32 & 12 & 256 & 1024 & 4 & 11.4M\\
% Base & ViT-B/32 & 12 & 384 & 1536 & 6 & 19.0M\\
% Large & ViT-B/32 & 12 & 512 & 2048 & 8 & 43.6M \\
% \bottomrule[0.95pt]
% \end{tabular}
% }
% \end{center}
% \end{table}

\begin{wraptable}{r}{0.5\textwidth}
\vspace{-13pt}
\centering
\setlength\tabcolsep{2pt}
\caption{\textbf{Details of PointSO model variants}. This table format follows~\citet{ViT21}.}
\vspace{-3pt}
\label{tab:PointSO_configs}
\resizebox{0.48\textwidth}{!}{
\begin{tabular}{lcccccc}
\toprule[0.95pt]
Model & CLIP & Layers & Hidden & MLP & Heads & \#Params\\
      &      &        & size   & size &       & \\
\midrule[0.6pt]
Small & ViT-B/32 & 12 & 256 & 1024 & 4 & 11.4M\\
Base  & ViT-B/32 & 12 & 384 & 1536 & 6 & 19.0M\\
Large & ViT-B/32 & 12 & 512 & 2048 & 8 & 43.6M\\
\bottomrule[0.95pt]
\end{tabular}
}
\vspace{-10pt}
\end{wraptable}

\subsection{PointSO Model Details}\label{app:pointso_details}
For PointSO, we utilize FPS + KNN to perform patchify and employ a small PointNet~\cite{PointNet17} as the patch encoder. Subsequently, a standard Transformer encoder is adopted as the backbone, followed by a single linear layer to map the output to a three-dimensional vector space. All parameter configurations follow prior work on point cloud representation learning~\cite{ACT23,ReCon23,ShapeLLM24}. Detailed hyperparameter and model configurations are provided in \cref{tab:hyper_params,tab:PointSO_configs}.

\subsection{SoFar-LLaVA Model Details}\label{app:model_details}
\begin{figure*}[t!]
\begin{center}
\includegraphics[width=1.0\linewidth]{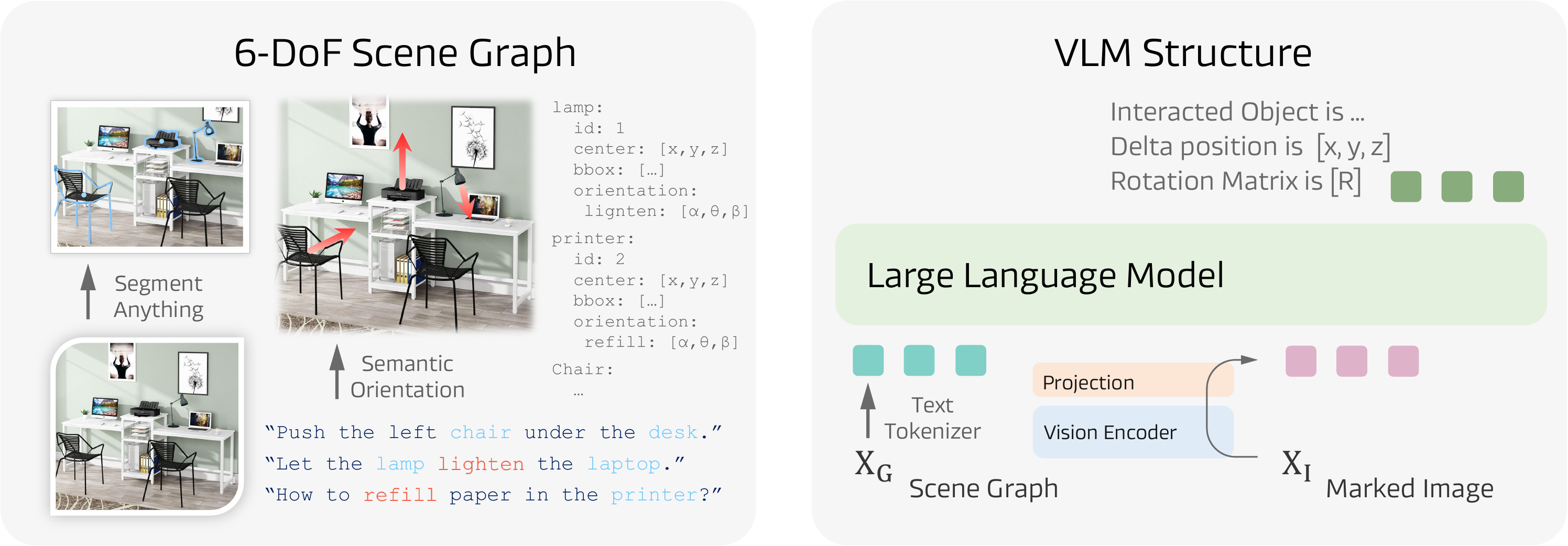}
\caption{\textbf{Pipeline of \sofar-LLaVA}, a fine-tuned VLM based on visual instruction tuning.
}
\label{fig:sofar_llava}
\vspace{-12pt}
\end{center}
\end{figure*}

In addition to leveraging the extensive knowledge and strong generalization capabilities of closed-source/open-source pretrained VLMs~\cite{ChatGPT22,gemini23,qwenvl23} for zero-shot or in-context learning, \sofar~can also enhance the planning performance of open-source models through visual instruction tuning for rapid fine-tuning. The pipeline of the model is illustrated in \cref{fig:sofar_llava}. A JSON-formatted 6-DoF scene graph, processed through a text tokenizer, along with the image refined by SoM~\cite{SoM23}, is fed into an LLM (\eg, LLaMA~\cite{LLaMA23,LLaMA2_23}) for supervised fine-tuning~\cite{LLaVA23}.
In the Open6DOR~\cite{Open6DOR24} task, we supplement the training dataset with additional samples retrieved and manually annotated from Objaverse~\cite{objaverse23}, ensuring alignment with the object categories in the original benchmark. This dataset includes approximately 3,000 6-DoF object manipulation instructions. Using this data, we construct dialogue-style training data based on ChatGPT and train the \sofar-LLaVA model. The training hyperparameters are detailed in \cref{tab:hyper_params}. Similarly, we finetune PointSO on this training dataset and achieve superior performance on the Open6DOR task.

\subsection{ChatGPT API Costs}
The knowledge of OrienText300K is derived from the annotations of 3D modelers on Sketchfab, combined with ChatGPT's filtering and comprehension capabilities. To generate semantic orientation annotations, we filter the 800K dataset of Objaverse~\cite{objaverse23} and apply ChatGPT to approximately 350K of the filtered data to generate semantic text-view index pairs. The OpenAI official API was used for these calls, with the GPT-4o version set to 2024-08-06 and the output format configured as JSON. The total cost for debugging and execution amounted to approximately \$10K.

\section{Additional Benchmark Statistic Analysis}
\subsection{6-DoF SpatialBench Analysis}
\begin{figure}[t!]
\begin{center}
\includegraphics[width=\linewidth]{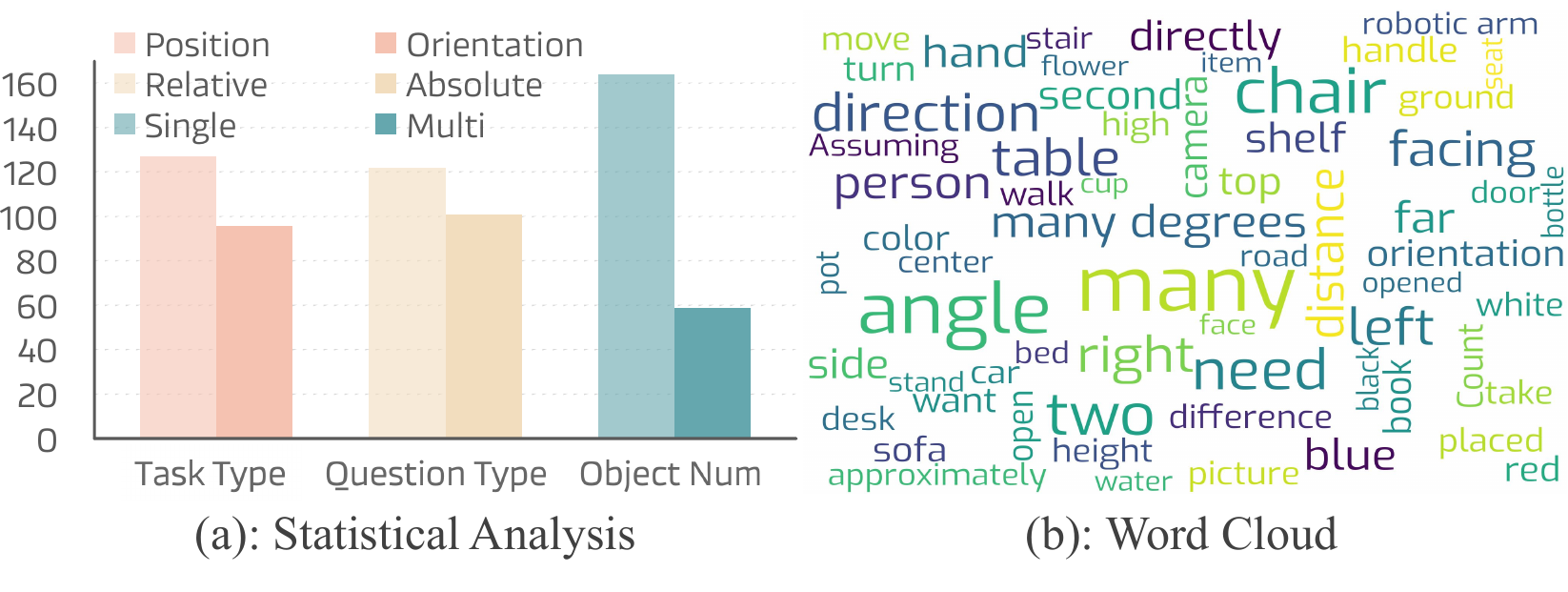}
\vspace{-20pt}
\caption{\textbf{6-DoF SpatialBench statistics}. (a) Statistical analysis of the task type, question type, and object relation. (b) Word cloud visualization.}
\label{fig:spatialvqa_statistic}
\end{center}
\end{figure}

We conduct a statistical analysis of the manually constructed 6-DoF SpatialBench, with category comparisons and word cloud visualizations shown in \cref{fig:spatialvqa_statistic}. We collect diverse image data from the internet, encompassing scenes such as indoor, outdoor, and natural landscapes. The questions may involve one or multiple objects, with varying levels of uncertainty in image resolution. Most importantly, we are the first to propose a VQA benchmark for orientation understanding, focusing on both quantitative and qualitative evaluation of orientation.

\subsection{Open6DOR V2 Analysis}
Open6DOR V2 builds upon Open6DOR V1 by removing some incorrectly labeled data, removing manual evaluation metrics, and integrating assets and metrics into Libero, enabling closed-loop policy evaluation. The detailed number of tasks is presented in \cref{tab:open6dorv2_statistic}, comprising over 4,500 tasks in total. Notably, we remove level 2 of the position track in Open6DOR V1~\cite{Open6DOR24} because it requires manual inspection, which is not conducive to open-source use and replication by the community. Besides, due to the randomness of object drops in the scene, approximately 8\% of the samples already satisfy the evaluation metrics in their initial state.

\section{Related Works}\label{app:related_work}
\subsection{Vision-Language Models for Spatial Understanding}
Vision-Language Models are rapidly being developed in the research community, driven by the storming lead in extending GPT-style~\cite{GPT1_18,GPT2_19,GPT3_20} Large Language Models (LLMs) like LLaMA~\cite{LLaMA23,LLaMA2_23} to VLMs~\cite{LLaVA23,LLaVA1.523,DreamLLM23,Emu23,LLaMA-Adapter23,ChatSpot23,omnispatial25}.
SpatialVLM~\cite{SpatialVLM24} pioneers this direction by constructing VQA data in spatial understanding from RGB-D, which is used for training an RGB-only VLM.
Following SpatialVLM, SpatialRGPT~\cite{SpatialRGPT24} extends RGB-based spatial understanding to RGB-D by constructing spatial understanding data using 3D scene graphs.
SpatialBot~\cite{SpatialBot24} explores RGB-D spatial reasoning through hierarchical depth-based reasoning.
Some other works propose visual prompting for improving GPT-4V's spatial understanding~\cite{3DAxiesPrompts23,SoM23,SpatialPIN24}.
Meanwhile, another line of works explores VLMs using 3D representations such as point clouds for 3D scene~\cite{3DLLM23,SceneLLM24} and object-centric~\cite{ShapeLLM24,pointllm23,GPT4Point24} understanding.
More recently, OmniSpatial~\cite{omnispatial25} proposed a comprehensive and challenging spatial reasoning benchmark.
Despite the remarkable progress, these works are limited to 3-DoF understanding, which is not actionable.
In contrast, we explore spatial understanding in 6-DoFs from RGB-D via VLMs.
Unlike vanilla 3D scene graphs used by SpatialRGPT for data construction, we propose orientation-aware 3D scene graphs realized by our proposed PointSO. 
In addition, we formulate spatial understanding as graph learning, where the scene graph nodes are directly input during inference.

\subsection{Language-Grounded Robot Manipulation}
Language-grounded robot Manipulation adopts the human language as a general instruction interface.
Existing works can be categorized into two groups:
\textbf{i)} \textit{End-to-end} models like RT-series~\cite{RT123,RT223,RTH24} built upon unified cross-modal Transformers with tokenized actions~\cite{PERACT22,InstructRL22,ALOHA23}, large vision-language-action models built from VLMs~\cite{OpenVLA24,dreamvla25}, or 3D representations~\cite{3DVLA24,RoboPoint24}.
Training on robot data such as Open X-Embodiment~\cite{OpenXEmbodiment24} and DROID~\cite{DROID24}, a remarkable process has been made.
However, the data \textit{scale} is still limited compared to in-the-wild data for training VLMs.
\textbf{ii)} \textit{Decoupled} high-level reasoning and low-level actions in large VLMs and small off-the-shelf policy models, primitives~\cite{SayCan22,CodeAsPolicy23,VoxPoser23,CoPa24,MOKA24,UniSim24,ManipAnywhere24,robomatrix24,dexvlg25,code_as_monitor25}, or articulated priors~\cite{a3vlm24,ManipLLM24}.
Our \sofar~lies in this group, where an open-world generalization property emerges from VLMs and our proposed PointSO is empowered by orientation-aware spatial understanding.

\subsection{3D Representation Learning}
Research on 3D Representation Learning encompasses various methods, including point-based~\cite{PointNet17,PointNet++17}, voxel-based~\cite{voxelnet15}, and multiview-based approaches~\cite{MVCNN3D15,MVTN21}. 
Point-based methods~\cite{PointNext22,PointTrans21} have gained prominence in object classification~\cite{ModelNet15,ScanObjectNN19} due to their sparsity yet geometry-informative representation. On the other hand, voxel-based methods~\cite{voxelrcnn21,SyncSpecCNN17,VPP23} offer dense representation and translation invariance, leading to a remarkable performance in object detection~\cite{ScanNet17} and segmentation~\cite{ShapeNetPart16,S3DIS16}.
The evolution of attention mechanisms~\cite{Transformer17} has also contributed to the development of effective representations for downstream tasks, as exemplified by the emergence of 3D Transformers~\cite{PointTrans21,groupfree21,voxeltransformer21}. Notably, 3D self-supervised representation learning has garnered significant attention in recent studies. PointContrast~\cite{PointContrast20} utilizes contrastive learning across different views to acquire discriminative 3D scene representations. Innovations such as Point-BERT~\cite{PointBERT22} and Point-MAE~\cite{PointMAE22} introduce masked modeling~\cite{MAE22,BERT21} pretraining into the 3D domain. 
ACT~\cite{ACT23} pioneers cross-modal geometry understanding through 2D or language foundation models such as CLIP~\cite{CLIP21} or BERT~\cite{BERT21}. 
Following ACT, {\scshape ReCon}~\cite{ReCon23} further proposes a learning paradigm that unifies generative and contrastive learning. 
PPT~\cite{ppt24} highlights the significance of positional encoding in 3D representation learning.
Additionally, leveraging foundation vision-language models like CLIP~\cite{ACT23,CLIP21} has spurred the exploration of a new direction in open-world 3D representation learning. This line of work seeks to extend the applicability and adaptability of 3D representations in diverse and open-world/vocabulary scenarios~\cite{OpenScene23,CLIPFO3D23,Lowis3D23,PLA23,PointGCC23}.

\section{Additional Discussions}
\subsection{Relation to Affordance \& 6-DoF Pose Estimation}
Conceptually, this semantic orientation is a counterpart of \textit{affordance}~\citep{Affordance77,AffordanceHRI16,MoveWithAffordanceMaps20,HandsAsAffordancesProbes22} but beyond,
as SO and affordance, all present potential actions and interactions with objects.
However, SO also contains the spatial understanding of intra-object part-level attributes more than affordance learning.
Compared to vanilla 6-DoF pose estimation, our proposed SO combined with the 3-DoF translation understanding, has the same DoF completeness.
The difference is, our proposed SO is grounded in languages, making it useful for open-world manipulation requiring complicated spatial reasoning~\cite{RobotsThatUseLanguage20,SayCan22,Open6DOR24}. 
In addition, our Semantic Orientation can be auto-labeled from Internet 3D data that achieves higher scalability, as introduced in \cref{sec:SO_def}.

\begin{figure*}[t!]
\includegraphics[width=\linewidth]{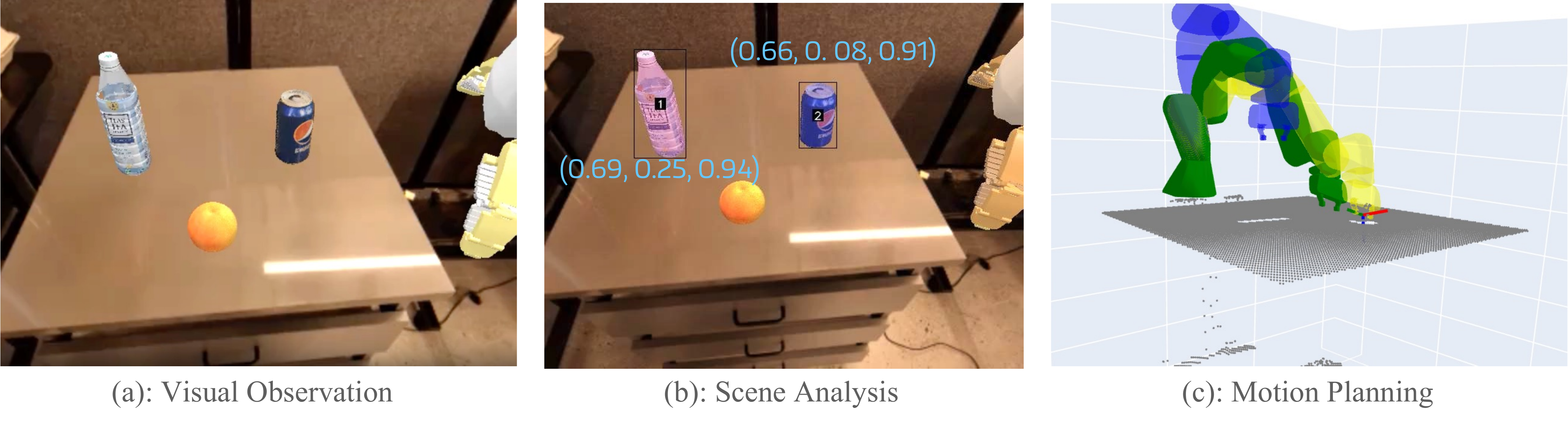}
\vspace{-18pt}
\caption{An example of \ours~how to finish ``move near'' task in SIMPLER~\cite{simplerenv24}.}
\label{fig:simpler_visual}
\vspace{-6pt}
\end{figure*}

\subsection{Comparison to Concurrent Works}
\noindent{\textbf{Comparison with ReKep~\cite{ReKep24}}}

Recently, ReKep has succeeded in executing complex robotic tasks, such as long-horizon manipulation, based on the relationships and constraints between spatial key points. 
Its structural design offers many insights that \sofar~can draw upon, yet it also presents several issues: 
(1) Overly customized prompt engineering. ReKep requires manually designed complex system prompts for each task during inference. 
While this approach may be described as ``no training'', it cannot be considered a true zero-shot transfer. In contrast, \sofar~achieves genuine zero-shot transfer by eliminating the need for any human involvement during inference; (2) Using constraints based solely on key points fails to capture the full 6-DoF pose integrity of objects. For example, in the ``pouring water'' task, merely bringing the spout of the kettle close to the cup may lead to incorrect solutions, such as the kettle overturning; (3) ReKep requires all key points to be present in the first frame, and each step of the process—from mask extraction to feature dimensionality reduction, clustering, and filtering—introduces additional hyperparameters.

\noindent{\textbf{Comparison with Orient Anything~\cite{orient_anything24}}}

Recently, Orient Anything also highlighted the importance of orientation in spatial perception and adopted a training data construction approach similar to Our PointSO. Our primary distinction lies in semantic orientation, which is language-conditioned orientation. In contrast, Orient Anything is limited to learning basic directions such as ``front'' and ``top''. By aligning with textual information, semantic orientation better enhances spatial perception, understanding, and robotic manipulation.

\subsection{Future Works}
Future work includes further expanding the OrienText300K with larger datasets like Objaverse-XL~\cite{ObjaverseXL23}, enhancing the performance of semantic orientation through self-supervised learning and pretraining methods~\cite{MAE22,CLIP21,ACT23,ReCon23}, and demonstrating its effectiveness in a broader range of robotic scenarios, such as navigation~\cite{GOAT24}, mobile manipulation~\cite{homerobot23}, lifelong learning~\cite{LIBERO23}, spatio-temporal reasoning~\cite{ReKep24,Leaf23,CrossVideoSC24,thinking24}, humanoid~\cite{OmniH2O24,SmoothHumanoidLCP24,Exbody24,humanup25}, and human-robot interaction~\cite{HOI4D22,InteractiveHO23}.

\begin{table*}[t!]
\centering
\caption{\textbf{Statistics of Open6DOR V2 Benchmark.} The entire benchmark comprises three independent tracks, each featuring diverse tasks with careful annotations. The tasks are divided into different levels based on instruction categories, with statistics demonstrated above.}
\vspace{-4pt}
\label{tab:open6dorv2_statistic}
\resizebox{1.0\linewidth}{!}{
\setlength{\tabcolsep}{3.5pt}
    \begin{tabular}{c|ccccc|cc|c|c|c|c|c}
    \toprule
        Track & \multicolumn{7}{c|}{Position-track} & \multicolumn{3}{c|}{Rotation-track} & 6-DoF-track & Totel\\
        \midrule
        Level  & \multicolumn{5}{c|}{Level 0} & \multicolumn{2}{c|}{Level 1}  & Level 0 & Level 1 & Level 2 & - & -\\
        \midrule
        Task Catog. &  Left & Right  & Top &  Behind &  Front & Between  & Center & Geometric & Directional & Semantic & - & - \\
        \midrule
        Task Stat. & 296 & 266 & 209 & 297 & 278 & 193 & 159 & 318 & 367 & 134 & 1810 & 4535\\
        
        \midrule
        Benchmark Stat. &\multicolumn{7}{c|}{1698} & \multicolumn{3}{c|}{1027} & 1810 & 4535\\
    \bottomrule 
    \end{tabular}
    \vspace{-16pt}
}
\end{table*}

\section{Additional Visualizations}\label{app:visualization}

\subsection{Robotic Manipulation}
As shown in \cref{fig:simpler_visual}, we present a visualization of executing a task named ``move near''.
According to the input image and task instruction - ``\textit{move blue plastic bottle near pepsi can}'', \ours~can predict the center coordinate of the target object (bottle) and relative target (pepsi can), and it would infer the place coordinate and produce a series of grasp poses.

\subsection{6-DoF SpatialBench}
\begin{figure*}[t!]
\centering
\includegraphics[width=1.0\linewidth]{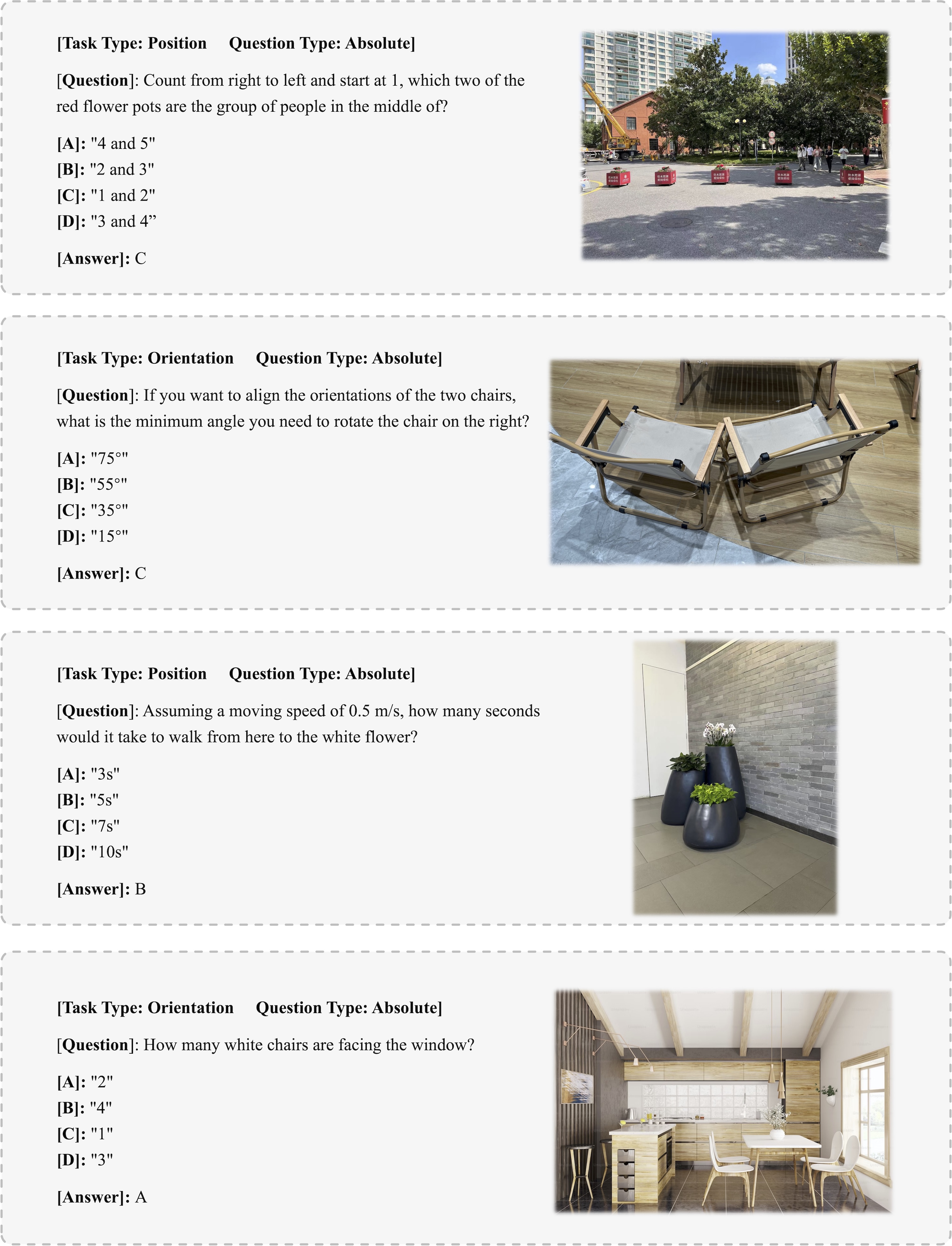}
\captionof{figure}{\textbf{Visualization example of 6-DoF SpatialBench data samples.}
% 6-DoF SpatialBench includes complex spatial reasoning of absolute numbers.
}
\label{fig:spatialbench_show1}
\end{figure*}

\begin{figure*}[t!]
\centering
\includegraphics[width=1.0\linewidth]{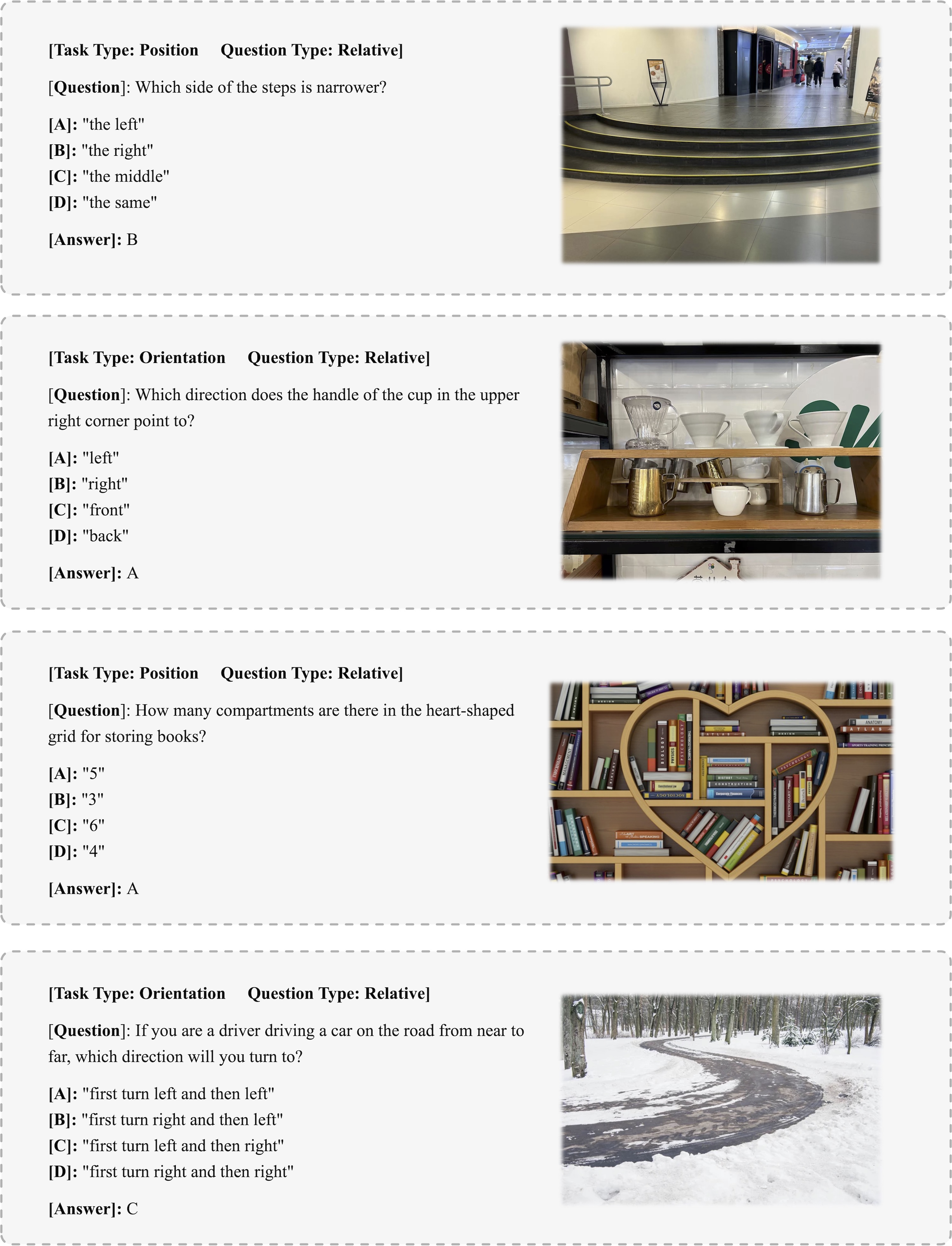}
\captionof{figure}{\textbf{Visualization example of 6-DoF SpatialBench data samples.}}
\label{fig:spatialbench_show2}
\end{figure*}

To further evaluate 6-DoF spatial understanding, we construct a 6-DoF SpatialBench.
We present examples of question-answer pairs from the 6-DoF SpatialBench, with quantitative and qualitative questions shown in \cref{fig:spatialbench_show1,fig:spatialbench_show2}, respectively. 
The benchmark we constructed is both challenging and practical, potentially involving calculations based on the laws of motion, such as ``\textit{Assuming a moving speed of 0.5 m/s, how many seconds would it take to walk from here to the white flower?}'' Moreover, it covers a wide range of spatially relevant scenarios across both indoor and outdoor environments.

\subsection{System Prompts}
\begin{figure*}[t!]
\centering
\includegraphics[width=1.0\linewidth]{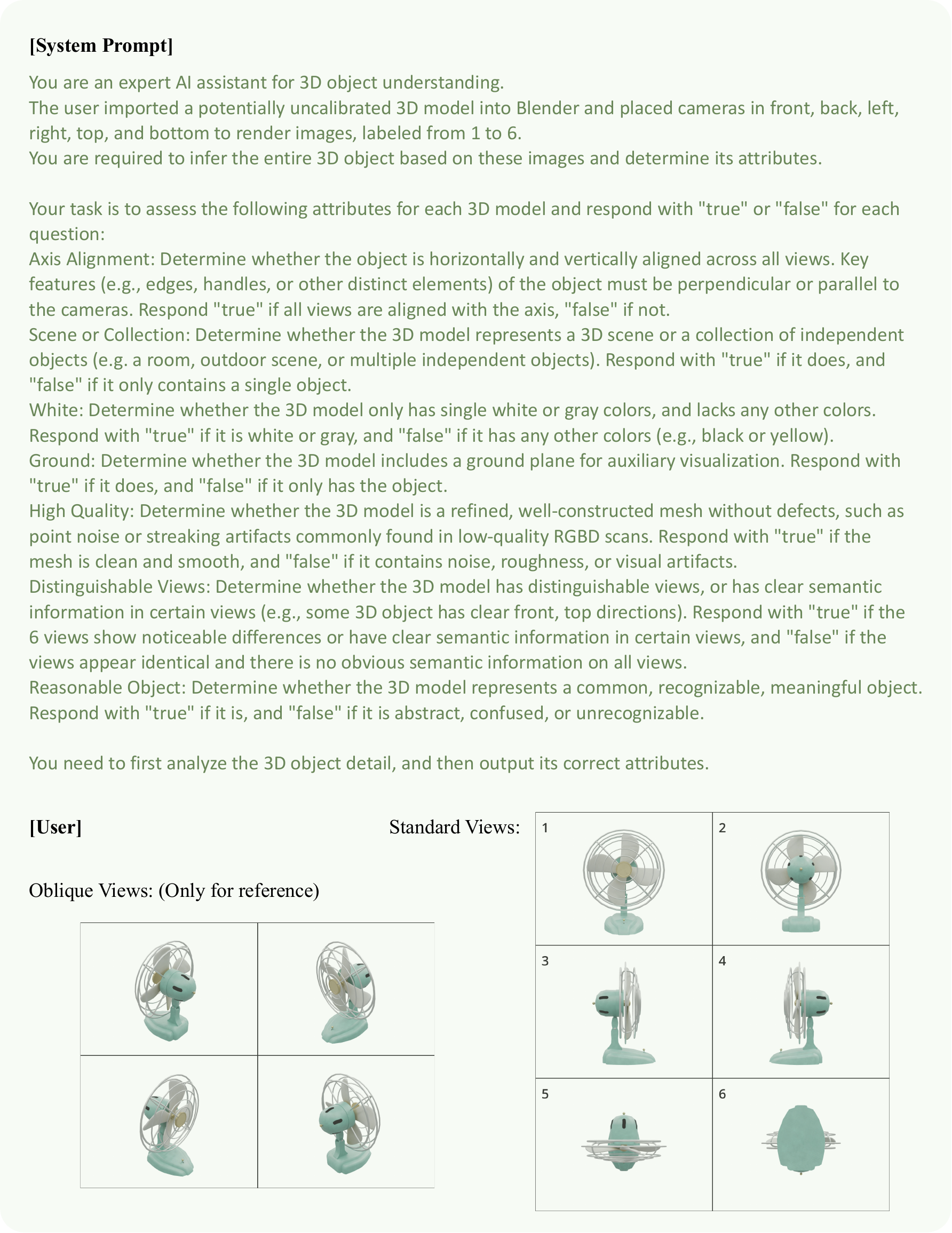}
\captionof{figure}{\textbf{The system prompt of GPT-4o used for filtering Objaverse data.}}
\label{fig:filter_prompt}
\end{figure*}

\begin{figure*}[t!]
\centering
\includegraphics[width=1.0\linewidth]{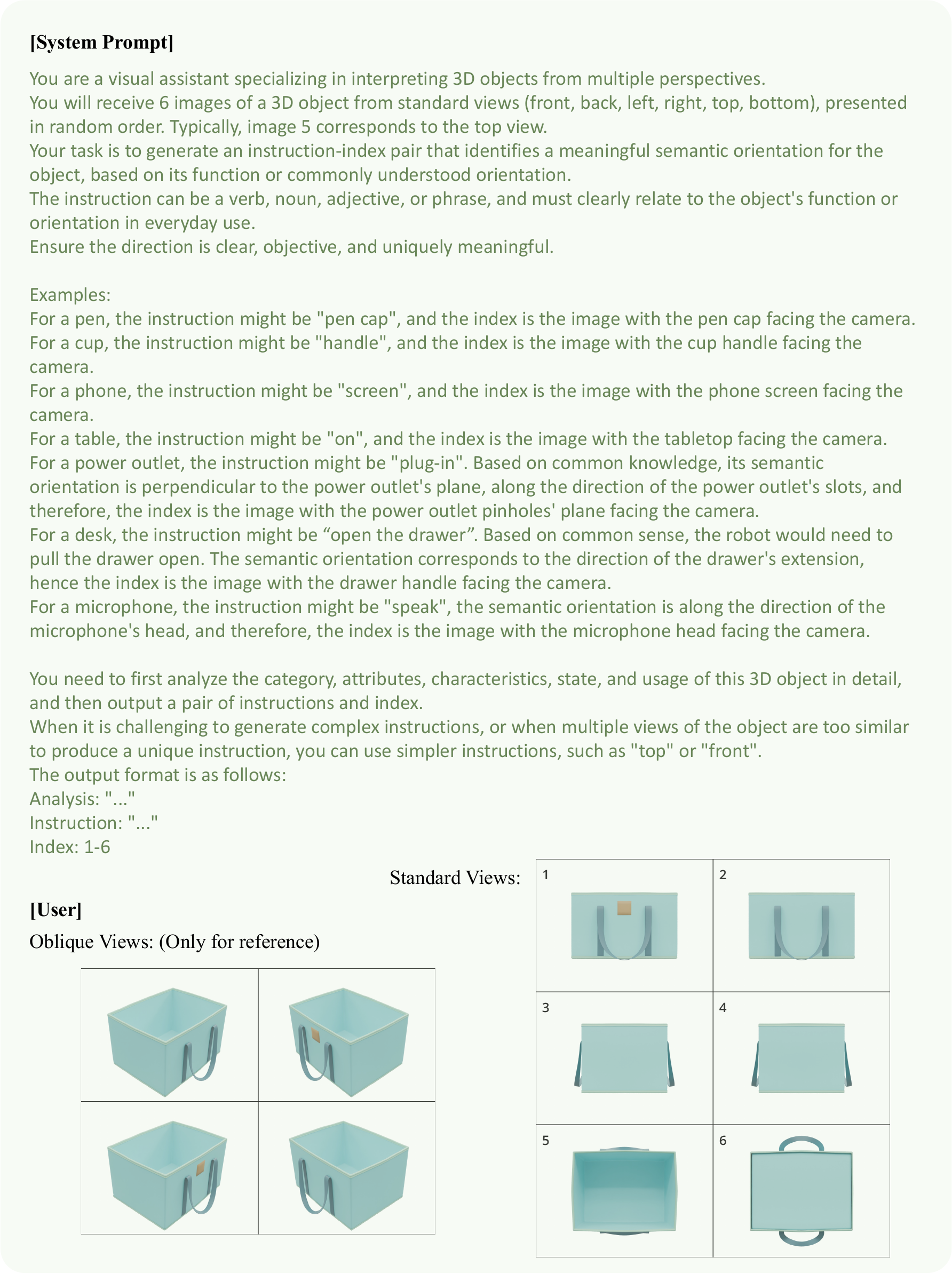}
\captionof{figure}{\textbf{The system prompt of GPT-4o used for generating semantic orientation-Index pairs.}}
\label{fig:instruction_prompt}
\end{figure*}

\begin{figure*}[t!]
\centering
\includegraphics[width=1.0\linewidth]{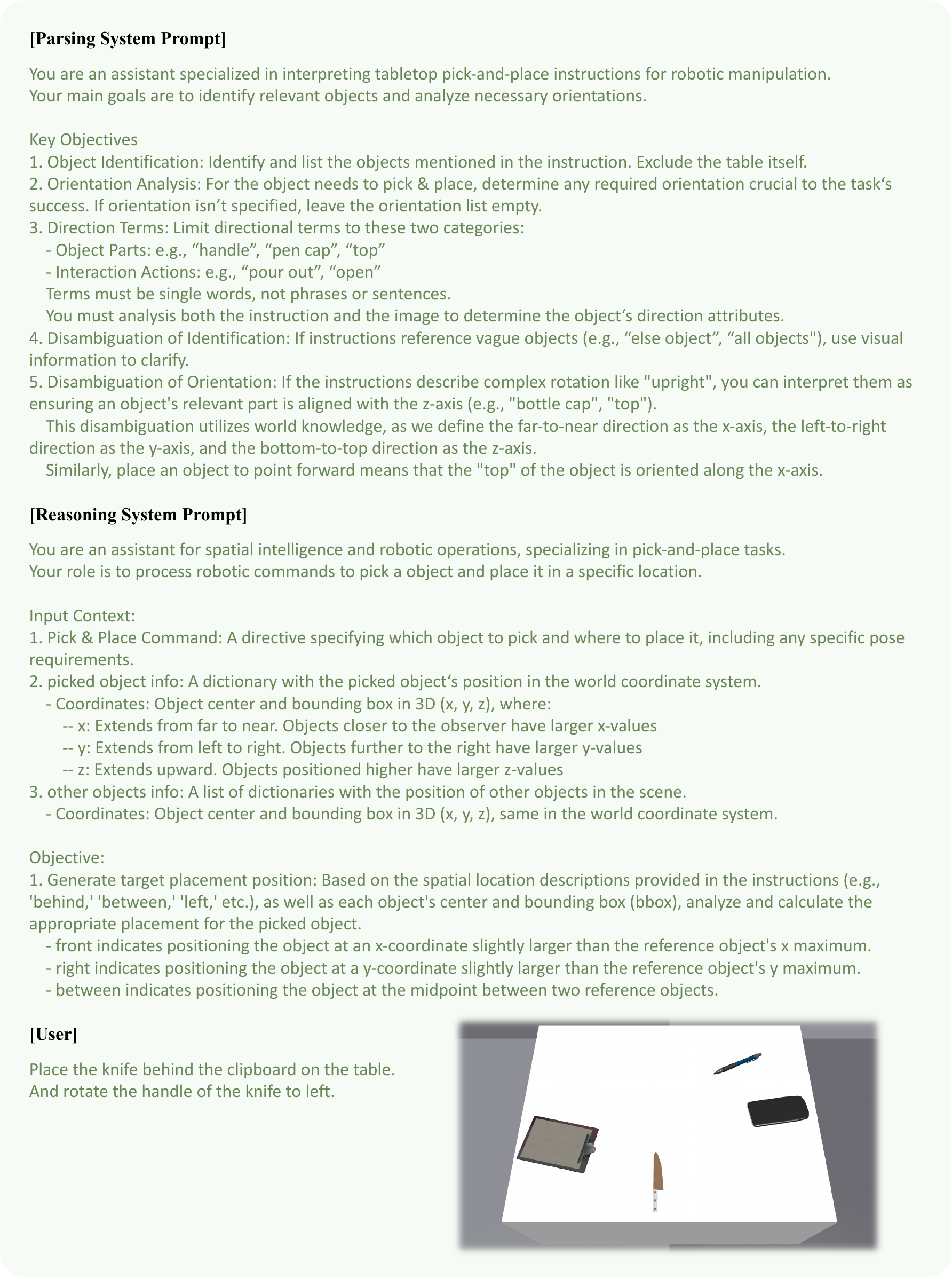}
\captionof{figure}{\textbf{The system prompt of Open6DOR tasks.}}
\label{fig:open6dor_prompt}
\end{figure*}

\begin{figure*}[t!]
\centering
\includegraphics[width=1.0\linewidth]{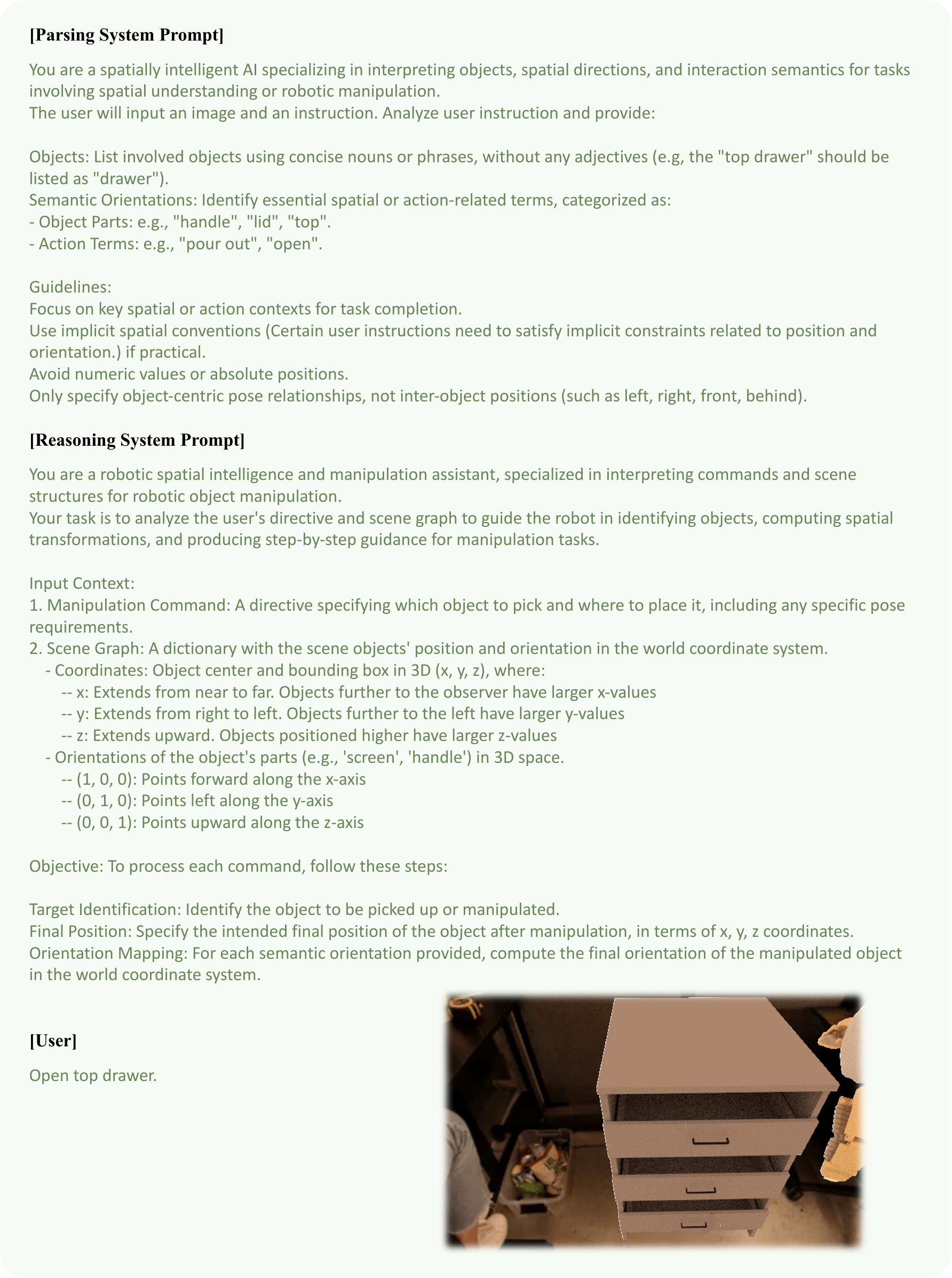}
\captionof{figure}{\textbf{The system prompt of general manipulation tasks.}}
\label{fig:manip_prompt}
\end{figure*}

\begin{figure*}[t!]
\centering
\includegraphics[width=1.0\linewidth]{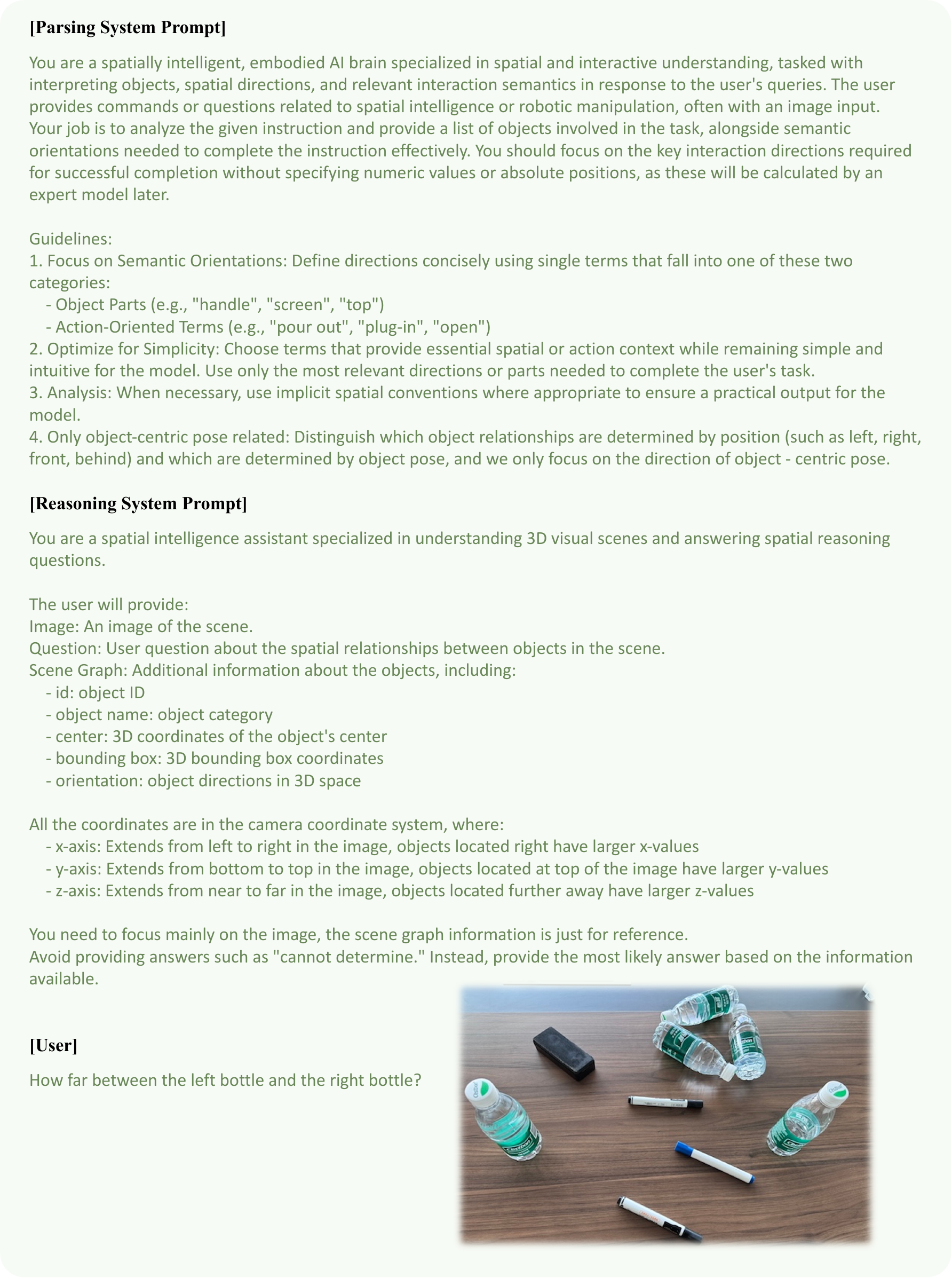}
\captionof{figure}{\textbf{The system prompt of visual-question-answering tasks.}}
\label{fig:vqa_prompt}
\end{figure*}

Prompt engineering significantly enhances ChatGPT's capabilities. The model's understanding and reasoning abilities can be greatly improved by leveraging techniques such as Chain-of-Thought~\cite{CoT22} and In-Context Learning~\cite{GPT3_20}. \cref{fig:filter_prompt,fig:instruction_prompt} illustrate the system prompt we used in constructing OrienText300K.
\cref{fig:open6dor_prompt}, \cref{fig:manip_prompt}, and \cref{fig:vqa_prompt} illustrate the system prompt we used when evaluating \sofar on Open6DOR (simulation), object manipulation (both simulation and real worlds), and VQA, respectively.
Note that different from previous methods~\cite{VoxPoser23,ReKep24}, \sofar does not require complicated in-context examples.

\section{Broader impacts}
\label{app:broader_impacts}
Our work on semantic orientation significantly enhances robotic spatial reasoning and manipulation capabilities, enabling more intuitive human-robot interaction. This advancement can improve efficiency in various industries, such as manufacturing and healthcare, and enhance the quality of life by assisting in tasks like elderly care and home automation. Additionally, it contributes to the broader field of AI research by providing new tools and benchmarks for spatial reasoning and language-grounded manipulation.

%%%%%%%%%%%%%%%%%%%%%%%%%%%%%%%%%%%%%%%%%%%%%%%%%%%%%%%%%%%%

\clearpage
\section*{NeurIPS Paper Checklist}

\begin{enumerate}

\item {\bf Claims}
    \item[] Question: Do the main claims made in the abstract and introduction accurately reflect the paper's contributions and scope?
    \item[] Answer: \answerYes{} % Replace by \answerYes{}, \answerNo{}, or \answerNA{}.
    \item[] Justification: We state the contributions in the abstract and introduction.

\item {\bf Limitations}
    \item[] Question: Does the paper discuss the limitations of the work performed by the authors?
    \item[] Answer: \answerYes{} % Replace by \answerYes{}, \answerNo{}, or \answerNA{}.
    \item[] Justification: We discuss the limitations at \cref{sec:limitation}.

\item {\bf Theory assumptions and proofs}
    \item[] Question: For each theoretical result, does the paper provide the full set of assumptions and a complete (and correct) proof?
    \item[] Answer: \answerNA{} % Replace by \answerYes{}, \answerNo{}, or \answerNA{}.
    \item[] Justification: N/A

    \item {\bf Experimental result reproducibility}
    \item[] Question: Does the paper fully disclose all the information needed to reproduce the main experimental results of the paper to the extent that it affects the main claims and/or conclusions of the paper (regardless of whether the code and data are provided or not)?
    \item[] Answer: \answerYes{} % Replace by \answerYes{}, \answerNo{}, or \answerNA{}.
    \item[] Justification: We provide detailed experiment information in the \cref{app:implementation_details} for reproduction.

\item {\bf Open access to data and code}
    \item[] Question: Does the paper provide open access to the data and code, with sufficient instructions to faithfully reproduce the main experimental results, as described in supplemental material?
    \item[] Answer: \answerYes{} % Replace by \answerYes{}, \answerNo{}, or \answerNA{}.
    \item[] Justification: We provide the data and code in the supplemental material.

\item {\bf Experimental setting/details}
    \item[] Question: Does the paper specify all the training and test details (e.g., data splits, hyperparameters, how they were chosen, type of optimizer, etc.) necessary to understand the results?
    \item[] Answer: \answerYes{} % Replace by \answerYes{}, \answerNo{}, or \answerNA{}.
    \item[] Justification: We provide detailed hyperparameters in the \cref{app:implementation_details} for reproduction.

\item {\bf Experiment statistical significance}
    \item[] Question: Does the paper report error bars suitably and correctly defined or other appropriate information about the statistical significance of the experiments?
    \item[] Answer: \answerYes{} % Replace by \answerYes{}, \answerNo{}, or \answerNA{}.
    \item[] Justification: In the real-world experiment depicted in \cref{fig:real}, we report error bars.

\item {\bf Experiments compute resources}
    \item[] Question: For each experiment, does the paper provide sufficient information on the computer resources (type of compute workers, memory, time of execution) needed to reproduce the experiments?
    \item[] Answer: \answerYes{} % Replace by \answerYes{}, \answerNo{}, or \answerNA{}.
    \item[] Justification: We provide detailed computer resources in the \cref{app:implementation_details}.
    
\item {\bf Code of ethics}
    \item[] Question: Does the research conducted in the paper conform, in every respect, with the NeurIPS Code of Ethics \url{https://neurips.cc/public/EthicsGuidelines}?
    \item[] Answer: \answerYes{} % Replace by \answerYes{}, \answerNo{}, or \answerNA{}.
    \item[] Justification: The paper has ensured anonymity and ethical standards.

\item {\bf Broader impacts}
    \item[] Question: Does the paper discuss both potential positive societal impacts and negative societal impacts of the work performed?
    \item[] Answer: \answerYes{} % Replace by \answerYes{}, \answerNo{}, or \answerNA{}.
    \item[] Justification: We discuss potential positive societal impacts and negative societal impacts in \cref{app:broader_impacts}.
    
\item {\bf Safeguards}
    \item[] Question: Does the paper describe safeguards that have been put in place for responsible release of data or models that have a high risk for misuse (e.g., pretrained language models, image generators, or scraped datasets)?
    \item[] Answer: \answerNA{} % Replace by \answerYes{}, \answerNo{}, or \answerNA{}.
    \item[] Justification: N/A

\item {\bf Licenses for existing assets}
    \item[] Question: Are the creators or original owners of assets (e.g., code, data, models), used in the paper, properly credited and are the license and terms of use explicitly mentioned and properly respected?
    \item[] Answer: \answerYes{} % Replace by \answerYes{}, \answerNo{}, or \answerNA{}.
    \item[] Justification: In \cref{sec:exp}, we properly credit all the public baselines and datasets utilized in this paper.

\item {\bf New assets}
    \item[] Question: Are new assets introduced in the paper well documented and is the documentation provided alongside the assets?
    \item[] Answer: \answerYes{} % Replace by \answerYes{}, \answerNo{}, or \answerNA{}.
    \item[] Justification: We include the documentation in the supplementary materials.

\item {\bf Crowdsourcing and research with human subjects}
    \item[] Question: For crowdsourcing experiments and research with human subjects, does the paper include the full text of instructions given to participants and screenshots, if applicable, as well as details about compensation (if any)? 
    \item[] Answer: \answerNA{} % Replace by \answerYes{}, \answerNo{}, or \answerNA{}.
    \item[] Justification: N/A

\item {\bf Institutional review board (IRB) approvals or equivalent for research with human subjects}
    \item[] Question: Does the paper describe potential risks incurred by study participants, whether such risks were disclosed to the subjects, and whether Institutional Review Board (IRB) approvals (or an equivalent approval/review based on the requirements of your country or institution) were obtained?
    \item[] Answer: \answerNA{} % Replace by \answerYes{}, \answerNo{}, or \answerNA{}.
    \item[] Justification: N/A

\item {\bf Declaration of LLM usage}
    \item[] Question: Does the paper describe the usage of LLMs if it is an important, original, or non-standard component of the core methods in this research? Note that if the LLM is used only for writing, editing, or formatting purposes and does not impact the core methodology, scientific rigorousness, or originality of the research, declaration is not required.
    %this research? 
    \item[] Answer: \answerNA{} % Replace by \answerYes{}, \answerNo{}, or \answerNA{}.
    \item[] Justification: N/A

\end{enumerate}

\end{document}